\documentclass{article}

\usepackage{arxiv}
\usepackage[utf8]{inputenc} 
\usepackage[T1]{fontenc}    
\usepackage{url}            
\usepackage{booktabs}       
\usepackage{amsmath, amsfonts, amssymb}       
\usepackage{nicefrac}       
\usepackage{microtype}      
\usepackage{lipsum}
\usepackage{graphicx}
\usepackage[bottom]{footmisc}
\usepackage[english]{babel}
\usepackage{caption}
\usepackage[normalem]{ulem}
\usepackage{color,soul}
\usepackage{authblk}
\usepackage{svg}
\usepackage{enumitem}
\usepackage{subcaption}
\newcommand{\beginsupplement}{%
        \setcounter{table}{0}
        \renewcommand{\thetable}{S\arabic{table}}%
        \setcounter{figure}{0}
        \renewcommand{\thefigure}{S\arabic{figure}}%
        \setcounter{section}{0}
        \renewcommand{\thesection}{Supplementary Section S\arabic{section}}%
     }

\usepackage{caption,graphicx,newfloat}
\usepackage[font=scriptsize,labelfont=bf]{caption}
\DeclareCaptionType{InfoBox}

\usepackage{verbatim}
\usepackage{algorithmic}
\usepackage{multicol, blindtext}
\usepackage{textcomp}

\def\BibTeX{{\rm B\kern-.05em{\sc i\kern-.025em b}\kern-.08em
    T\kern-.1667em\lower.7ex\hbox{E}\kern-.125emX}}

\usepackage{lmodern}
\usepackage[T1]{fontenc}
\usepackage{float}

\usepackage{hyperref, url}
\usepackage{xr}

\makeatletter
\newcommand*{\addFileDependency}[1]{
  \typeout{(#1)}
  \@addtofilelist{#1}
  \IfFileExists{#1}{}{\typeout{No file #1.}}
}
\makeatother

\begin{document}

{
\textbf\newline \title{\Large \emph{Traditional} machine learning vs. \emph{deep} learning from dynamic graph representations of proteins' 3D folds in the task of protein structure classification}
}

\vspace{-1cm}

\author{Aydin Wells, Francis A. Gatsi, Aaron Striegel, and Tijana Milenkovi\'{c}$^{*}$}
\date{}

\maketitle
\begin{center}
\vspace{-0.8cm}

Department of Computer Science and Engineering, University of Notre Dame, USA\\
$^*$Corresponding author (email: tmilenko@nd.edu)\\

\end{center}

\bigskip

\begin{abstract}
Protein structure classification (PSC) uses supervised learning to predict a protein's CATH/SCOP(e) class from the protein's sequence or 3D structural feature(s). We already modeled 3D structures as (static) protein structure networks (PSNs), demonstrating the competitiveness of PSN-based features to sequence or direct (i.e. non-network) 3D structural features in the PSC task. More recently, we demonstrated the power of features extracted from dynamic PSNs over features extracted from static PSNs (and thus by transitivity over sequence and direct 3D structural features) in the same task. That dynamic PSN approach used traditional machine learning (ML), combining manual (pre-engineered) features with an off-the-shelf classifier. Here, we evaluate whether 
automatic deep learning (DL) from the dynamic PSNs yields improvements. 
Our evaluation on 72 datasets spanning $\sim$44,000 CATH- or SCOPe-labeled dynamic PSNs reveals that in terms of PSC accuracy, traditional ML and DL are (close to) tied for a large majority of the datasets, while DL is on average 10+ times slower. We are the first to evaluate traditional ML vs. DL in the dynamic PSN-based PSC task.
\end{abstract}

\section{Introduction}\label{sect:introduction}

Determining protein function \emph{experimentally} is resource-consuming. This 
has motivated \emph{computational} prediction of protein function \cite{zhou2019cafa}. A protein’s amino acid sequence determines its 3D structure/fold, which in turn determines its function \cite{whitford2013proteins}. Consequently, computational methods extract features from sequences or 3D structures to predict functions. Before such features are used for function prediction, it is desirable to first validate that the features actually meaningfully reflect the proteins' folds. Protein structure classification (PSC)   \cite{newaz2020network} serves exactly this purpose. PSC is a task of predicting a protein' structural class based on the protein's feature(s) via supervised classification; the structural classes originate from the CATH Protein Structure Classification database \cite{greene2007cath} or the Structural Classification of Proteins (SCOP) or SCOP-extended (SCOPe) databases \cite{chandonia2019scope,murzin1995scop}. 
Note that there exists a related task of  protein structure comparison, which quantifies similarities between proteins based on their features (often in unsupervised fashion) and then examines whether the highly-similar proteins belong to the same CATH/SCOP(e) classes \cite{faisal2017grafene,vanKempen2024Foldseek}. Despite the related application focus of the PSC and structure comparison tasks, the two are different computationally and evaluation-wise \cite{Chen2026protein,faisal2017grafene,newaz2020network}. The focus of this paper is on the task of PSC.  

\begin{figure*}[h]
        \begin{center}
    \includegraphics[width=1\textwidth,trim= 0.9cm 17.3cm 0.9cm 0cm]{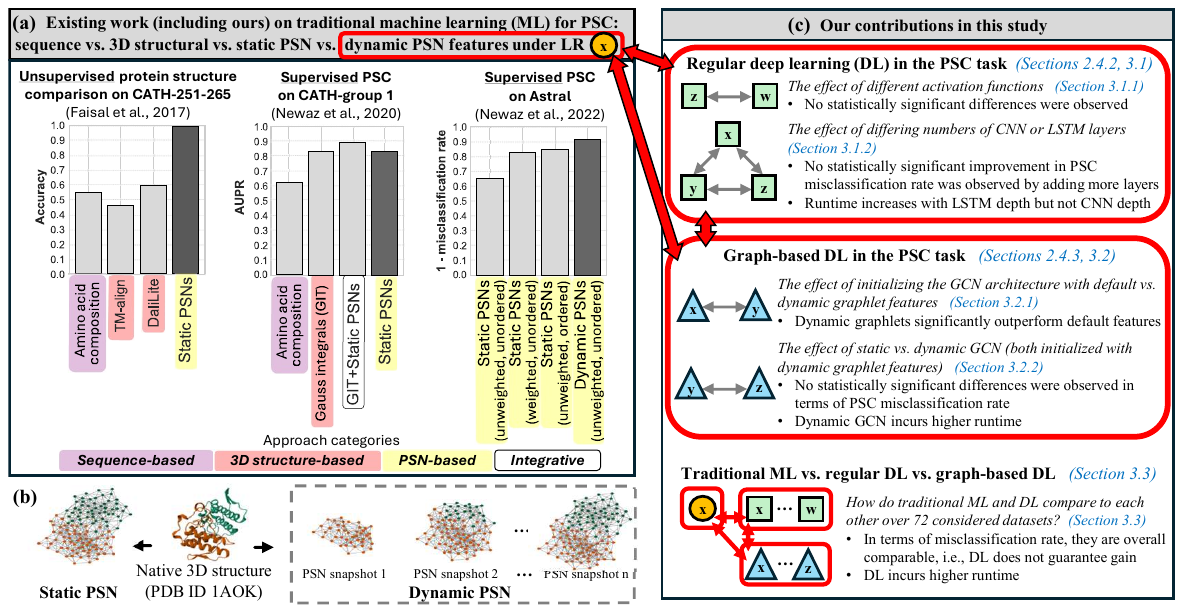}
        \end{center}
        \vspace{-0.3cm}
        \caption{Summary of our study. \textbf{(a)} Relevant prior work demonstrating the competitiveness of static PSNs to sequence and direct (non-network) 3D structural approaches in the task of protein structure comparison (left) \cite{faisal2017grafene} and our considered PSC task (middle) \cite{newaz2020network}, as well as the power of dynamic over static PSNs in the PSC task (right) \cite{newaz2022multi}. This existing dynamic PSN approach combining dynamic graphlet features and LR falls under the traditional ML paradigm. In each bar plot, the dark grey bar is the PSN approach whose power we are illustrating. In the right-most bar plot, the most fair comparison is between dynamic PSNs and the first static PSN model, as both are unweighted and unordered; the other two static PSN models are weighted or ordered. Each of the three bar plots contains representative results on a particular dataset (shown in the bar plot title) and for a particular performance measure ($y$-axis; the higher the value, the better). For full results (all considered datasets and performance measures), which align with the illustrated results, see the publications shown in the bar plot titles \cite{faisal2017grafene,newaz2020network,newaz2022multi}. \textbf{(b)} Illustration of the static and  dynamic PSN of the native 3D structure for the same protein. \textbf{(c)} Our contributions in this study: evaluation of traditional ML vs. DL in the task of dynamic PSN-based PSC. The regular DL paradigm (top) applies to the CNN+LSTM architecture and its variants. The graph-based DL paradigm (middle) applies to the GCN architecture and its variants. We compare the overall best variant from each DL paradigm to the existing traditional ML approach. Color-filled shapes (circles, squares, triangles) corresponds to the methods from Table \ref{tab:list_of_methods}.
        }
        \label{fig:overview}
    \end{figure*}

Historically, features for PSC have been extracted from sequences or 3D structures. More recently, features extracted from protein structure network (PSN) representations of 3D structures have emerged \cite{newaz2020network}. In a PSN, nodes are amino acids and edges capture their spatial proximity within the 3D fold \cite{faisal2017grafene,newaz2020network,newaz2020codon}. PSNs enable the extraction of network features that could not be directly extracted from sequences or 3D structures. PSNs are effective in the PSC task, achieving accuracy comparable to, or exceeding, that of sequence or direct (non-network) 3D structural approaches (Fig. \ref{fig:overview}(a)), while improving or maintaining competitive runtimes \cite{newaz2020network}. Similarly, the effectiveness of PSNs was also shown in the task of protein structure comparison \cite{faisal2017grafene,malod2014gr}, where static PSNs outperformed e.g. TM-align \cite{Zhang2005TMalign} (Fig. \ref{fig:overview}(a)), as well as in the task of protein function prediction \cite{gligorijevic2021structure}.  

These earlier successes were based on \textit{static} PSNs, where the final, native 3D structure is modeled as a single PSN (Fig. \ref{fig:overview}(b)). More recently, we introduced a \textit{dynamic} PSN as a ``proxy'' for mimicking a protein's co-translational folding dynamics, by capturing a series of progressively increasing ``proxy'' intermediates along the protein's folding pathway \cite{newaz2022multi}. A protein's first ``proxy'' intermediate contains the first $k$ amino acids of the entire sequence and the corresponding substructure of the native 3D structure, the second ``proxy'' intermediate  contains the first $2k$ amino acids of the entire sequence and the corresponding substructure of the native structure, etc; this continues until arriving at the last ``proxy'' intermediate, which captures the entire sequence and the whole native 3D structure \cite{newaz2022multi}; $k=5$ is recommended \cite{wells2026unavailability}. Each ``proxy'' intermediate is modeled as a (static) PSN snapshot, where an edge connects two amino acids in a given snapshot if the spatial distance between their C$\alpha$ atoms is within $6\mathring{A}$ in the native 3D structure. The collection of all snapshots is the protein's dynamic PSN (Fig. \ref{fig:overview}(b)). Note that an alternative term for a dynamic PSN is a multi-layer sequential PSN \cite{newaz2022multi}. Also, note that just like a static PSN, a dynamic PSN is derived from the native 3D structure, because the data on 3D structures of the actual protein folding intermediates are unavailable \cite{wells2026unavailability}; hence the term ``proxy''.

In that existing original dynamic PSN study \cite{newaz2022multi}, we rigorously evaluated dynamic vs. static PSNs in the PSC task \cite{newaz2022multi} on 72 datasets containing 34,630 CATH and 9,329 SCOPe domains with available experimentally determined 3D structures in the Protein Data Bank (PDB) \cite{wwpdb2019protein}; the datasets span all four hierarchy levels of CATH and SCOPe. We evaluated the two PSN types in a controlled manner: we used the same PSN feature type (the only difference being static vs. dynamic), the same machine learning (ML) classifier, and identical training and testing splits for classification via cross-validation. We measured misclassification rate -- the portion of protein domains in the test data whose predicted classes do \emph{not} match their ground-truth CATH/SCOPe classes, expressed in the [0,1] range; the lower the misclassification rate, the better.
Dynamic PSNs were significantly superior  (adjusted $p$-value, i.e. $q$-value, $\leq 10^{-6}$) to static PSNs (Fig. \ref{fig:overview}(a)) \cite{newaz2022multi}. Thus, by transitivity, dynamic PSNs were also superior to the existing sequence and direct (non-network) 3D structural approaches that static PSNs were already shown to be competitive to \cite{faisal2017grafene,newaz2020network}, per the above discussion and as illustrated in Fig. \ref{fig:overview}(a). Importantly, dynamic PSNs achieved low misclassification rates of under 0.1 on 90\% of the 72 considered datasets
\cite{newaz2022multi}.

That existing dynamic PSN approach \cite{newaz2022multi},  a proof-of-concept for capturing ``proxy'' folding dynamics, is a ``traditional ML'' approach -- it uses manual,  pre-engineered features with an off-the-shelf classifier. The features are dynamic graphlets, Lego-like building blocks of networks \cite{hulovatyy2015exploring}. The classifier is logistic regression (LR). Note that an attempt to use a different off-the-shelf classifier in the task of PSN-based PSC did not yield gain compared to using LR \cite{newaz2020network}. 

Dynamic graphlet features of a protein are organized as a matrix whose dimension
varies across proteins due to their sequence length differences. However, off-the-shelf classifiers like LR require fixed-dimension features over all proteins. For compatibility, each dynamic graphlet matrix had to be user-transformed into a uniform-dimension feature vector via a heuristic prior to training. This can yield information loss or noise, as shown on static PSNs/static graphlets \cite{guo2019weighted}. Yet, even the preprocessed dynamic graphlet features under the simple LR  yielded a low PSC misclassification rate \cite{newaz2022multi}, likely due to the power of dynamic graphlets rather than LR.

In this study, we ask whether improvements over the existing dynamic PSN approach are possible. We seek to allow for learning from raw, variable-dimensional dynamic graphlet matrices without user interference \cite{guo2019weighted}. Motivated by this, and by the existing approach's suboptimal PSC performance on certain datasets \cite{newaz2022multi}, we examine whether deep learning (DL) can automatically extract additional signal from dynamic PSNs in the PSC task. This is because DL's capacity to automatically extract features from raw sequence or 3D structural inputs has already enabled breakthroughs in protein structure and function prediction as well as protein design \cite{dai2025survey}.
    
We explore two DL paradigms (Fig. \ref{fig:overview}(c)): (i) ``regular'' and (ii) ``graph-based''. These correspond to: (i) using dynamic PSN (graphlet) features only to initialize a DL architecture combining convolutional neural networks (CNNs) with bidirectional long short-term memory (LSTM) layers, but the architecture itself does not operate on the PSNs, and (ii) using a graph convolutional network (GCN) that explicitly operates on the PSNs while maybe also initialized by dynamic PSN features. Note that in the dynamic graphlet matrix of a PSN, nodes are ordered as they occur in the protein sequence; this is exactly why we can use regular DL on PSNs in the first place. For (i), we extend to dynamic PSNs a CNN+LSTM approach that was validated on static PSNs \cite{guo2019weighted}. For (ii), we extend to dynamic PSNs two prominent GCN approaches \cite{kipf2016semi,manessi2020dynamic}. For each DL paradigm, we consider multiple architecture variants (Table \ref{tab:list_of_methods}). We perform thorough and controlled PSC evaluation within each DL paradigm. Then, we compare the best variant from each paradigm to the existing, traditional ML dynamic PSN approach (plus several other ``baseline'' methods that are not based on dynamic PSNs -- see Section \ref{sect:results_compare_overall} for details). We are the first to evaluate traditional ML vs. DL in the dynamic PSN-based PSC task. 

\section{Materials and methods}\label{sect:materials_methods}

    \begin{table*}
        \begin{center}
        \includegraphics[width=1\textwidth]{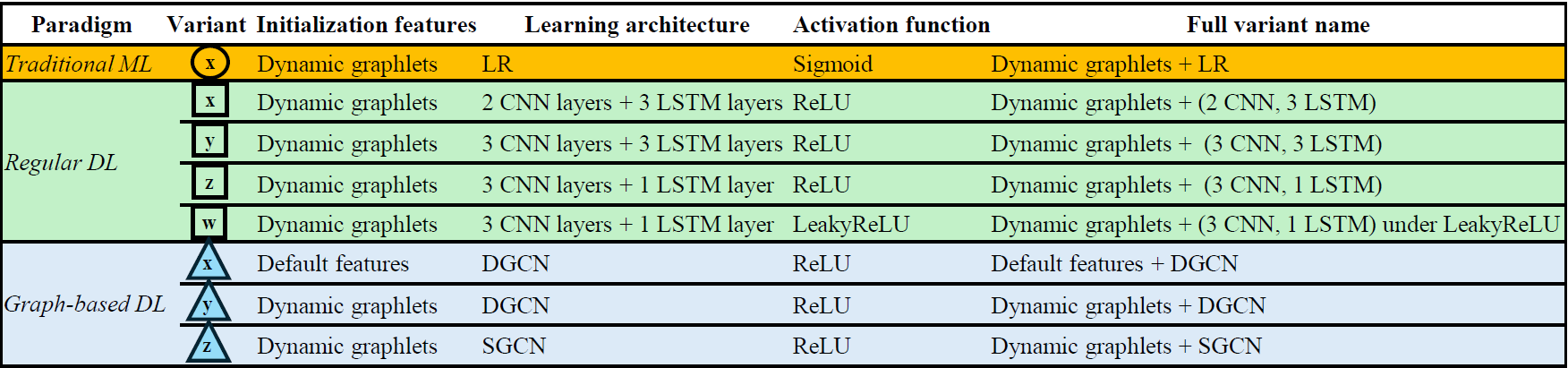}
        \end{center}
        \caption{Summary of the considered dynamic PSN-based methods. The column ``Variant'' is relevant only for Fig. \ref{fig:overview}. Note that we benchmark these methods against each other as well as against additional methods that are not based on dynamic PSNs.}
      \vspace{-0.3cm}
        \label{tab:list_of_methods}
    \end{table*}

\subsection{Data}\label{sect:data}

We evaluate on a comprehensive PSC benchmark comprising 72 datasets from the original dynamic PSN study \cite{newaz2022multi}, which contain 34,630 CATH and 9,329 SCOPe domains with available experimentally determined 3D structures in the PDB. The 72 different datasets span all four levels of the CATH and SCOPe, and as such vary substantially in the number of domains as well as the number of classes. This enables a rigorous and comprehensive PSC method evaluation.

Specifically, the datasets were created as follows \cite{newaz2022multi}. For CATH, one dataset was created at the first level that contains protein domains labeled by one of the three primary-level CATH classes (mainly $\alpha$, mainly $\beta$, and $\alpha/\beta$). For each of the three, a dataset was created at the second CATH level containing protein domains labeled by all subclasses of the given primary-level class; this yielded three datasets at the second CATH level. Following the same logic, 16 datasets were created at the third level, and 28 at the fourth (most fine-grained) level, resulting in 48 CATH datasets over all four CATH levels. Applying the same hierarchical construction procedure to SCOPe yielded an additional 22 datasets. Over CATH and SCOPe, these account for 70 datasets. Two additional low-sequence-identity benchmarks were constructed: the Astral dataset ($\leq 40\%$ pairwise sequence identity) and the Scop25\% dataset ($\leq 25\%$ sequence identity). This brings the  total to 72 datasets \cite{newaz2022multi}. 

For additional details, see \ref{sect:sup_data}.

\subsection{Dynamic PSNs}\label{sect:PSNs}

We construct a dynamic PSN as described in Section \ref{sect:introduction}, illustrated in Fig. \ref{fig:overview}(b), and described in full detail in our previous work \cite{newaz2022multi}.

\subsection{Dynamic PSN features -- dynamic graphlets}\label{sect:features}

For each dynamic PSN, we extract dynamic graphlets to characterize the change in network structure/topology over time (i.e. during the ``proxy'' folding dynamics; Section \ref{sect:introduction} and Fig. \ref{fig:overview}(b)). Graphlets are small, connected, induced subgraphs  -- such as a triangle or a square -- that capture  higher-order  organization of a network \cite{newaz2019graphlets,prvzulj2004modeling}. Graphlets have been extended from the static to dynamic context \cite{hulovatyy2015exploring}; dynamic graphlets allow for tracking how topological patterns appear, persist, or disappear in a network over time. 

\textcolor{black}{For each node in a dynamic network, its dynamic graphlet degree vector (dGDV) is computed, which counts the number of times that node participates in each dynamic graphlet automorphism orbit; we consider all graphlets on up to four nodes, with up to six temporal events, and use the ``dcount'' algorithm for dynamic graphlet counting  \cite{hulovatyy2015exploring}. Taking dGDVs for all nodes in a dynamic PSN, while ordering the nodes as they appear in the protein domain sequence, yields the dynamic dGDV matrix (dGDVM) for the entire network, i.e. for the corresponding protein domain. Given a domain with $n$ amino acids (i.e. its dynamic PSN with $n$ nodes), and given $o=3,727$ orbits for the considered dynamic graphlets, the dGDVM for that PSN has dimension $n \times o$; different domains/PSNs can have different numbers of amino acids/nodes.}
Because there are many orbits (dGDVM columns) for which the counts are zero across all dynamic PSNs over all datasets, as established \cite{newaz2022multi}, we remove all such columns. This results in 211 columns remaining in the non-zero dGDVMs. The 211-column dGDVMs, and in some cases also the dynamic PSNs themselves (i.e. their connectivity information) serve as the input into considered approaches (Fig. \ref{fig:fig_methods} and Section \ref{sect:classification_frameworks}).

\subsection{\textcolor{black}{Classification frameworks based on dynamic PSNs}}\label{sect:classification_frameworks}

The three classification frameworks/learning paradigms (traditional ML, regular DL, and graph-based DL) and their variants that we consider are listed in Table \ref{tab:list_of_methods}, illustrated in Fig. \ref{fig:fig_methods}, and discussed next.
    
\subsubsection{The existing LR (traditional ML) architecture}\label{sect:LR}

We employ the same LR–based methodological framework introduced and detailed in our previous study \cite{newaz2022multi}, as follows. 

LR requires fixed-length feature vectors as input.
With the $n \times 211$ non-zero dGDVMs produced in the previous steps (Section \ref{sect:features}), $n$ can vary across dynamic PSNs. LR cannot directly operate on these variable-size matrices. Instead, in the LR framework, it is necessary to obtain fixed-dimensional versions of the non-zero dGDVMs. This is achieved as follows. Each non-zero dGDVM is first transformed into a graphlet orbit correlation matrix (GCM) \cite{yaverouglu2014revealing}, which computes pairwise correlations between the orbits/columns of the dGDVM. As a result, each protein domain/dynamic PSN is now represented by the $211\times 211$ GCM. Next, the upper triangular portion of each GCM is flattened into a 1-dimensional feature vector of length ${211 \choose 2}$, yielding one such  vector per PSN. The vectors for all $i$ dynamic PSNs in a dataset are stacked into a matrix of size $i \times {211 \choose 2}$. Principal component analysis (PCA) is then applied to that matrix to reduce dimensionality while retaining 90\% of the variance, producing a transformed matrix of size $i \times d$, where $d$ is the number of principal components required to achieve this variance threshold. Each row of the resulting matrix represents a fixed-length dynamic PSN feature vector and is used as input to the LR classifier together with its corresponding CATH or SCOPe label.

For a given dataset containing $i$ protein domains (i.e. dynamic PSNs), each domain being labeled by a structural class, where the total number of unique structural classes over all $i$ domains is $c$, the LR framework adopts a one-vs-rest strategy to address the multi-class classification problem: $c$ independent LR models are trained, each corresponding to one of the $c$ structural classes. For each domain, the  $j^{\text{th}}$ classifier ($j=1,2,3,...,c$) is trained to distinguish a given domain belonging to class $j$ vs. to all the remaining classes. This enables the use of standard binary LR while extending it to a multi-class one. Model training, hyperparameter tuning (specifically L2 regularization strength), and performance evaluation are performed using five-fold cross-validation (Section \ref{sect:evaluation}). At the time of testing, all $i$ fixed-length dynamic PSN feature vectors (resulting from the multiple transformations including PCA) that correspond to the  $i$ domains/dynamic PSNs in a given dataset, are provided as input to all $c$ trained LR models. For each input vector (i.e. each domain), the LR models produce $c$ probability scores, one from each classifier. Then, the structural class assigned to (predicted for) a given domain is the class with the highest probability score out all of all $c$ probability scores. The $c$ LR models are implemented in Python using the \texttt{scikit-learn} library. 

\begin{figure*}[ht!]
    \begin{center}
    \includegraphics[width=0.86\textwidth,trim= 0.8cm 3.7cm 0.8cm 0cm]{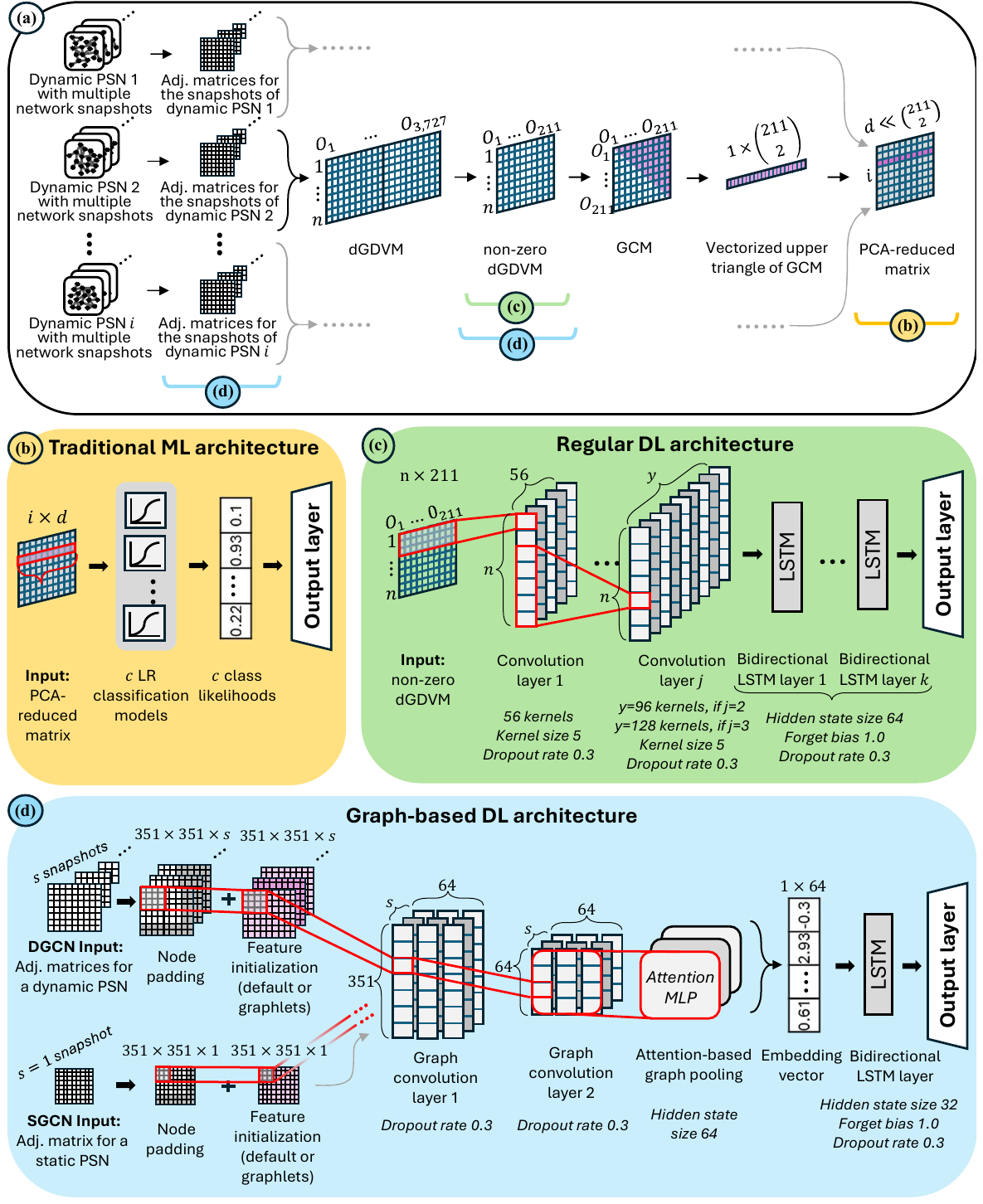}
    \end{center}
    \caption{Pipeline for \textbf{(a)} dynamic PSN generation, dGDVM feature extraction, and feature transformation (all matching our previous work \cite{newaz2022multi}). The resulting features from specific parts of this pipeline are used as input into the three considered learning paradigms (denoted with a yellow, green or blue colored bracket and a colored circle -- where each circle contains a letter corresponding to a paradigm/panel in the figure).
    \textbf{(b)} A schematic of the traditional ML architecture that we use in our study (matching our previous work \cite{newaz2022multi}).
    \textbf{(c)} A schematic of the regular DL architecture that is inspired by and adapted from  previous work on static PSNs \cite{guo2019weighted}.
    \textbf{(d)} A schematic of the graph-based DL architecture (for both its DGCN and SGCN variants)  \cite{manessi2020dynamic} that we use.
    }
    \label{fig:fig_methods}
\end{figure*}

We illustrate the LR (i.e. traditional ML) framework in Fig. \ref{fig:fig_methods}(b). For additional details on it, see our previous study \cite{newaz2022multi}.

\subsubsection{The regular DL architecture and its variants}\label{sect:CNN_LSTM}

Motivated by prior work demonstrating the effectiveness of combining (weighted) static graphlets/PSNs with deep neural networks in the PSC task \cite{guo2019weighted}, we extend this idea to dynamic graphlets/PSNs. Unlike the LR framework described above, which requires transforming the non-zero dGDVM of a protein domain into a fixed-length feature vector prior to classification, DL models can automatically operate directly on a variable-length feature input. Consequently, DL models can process the raw non-zero dGDVM without any transformations. In this study, one of the DL paradigms that we consider is what we refer to as ``regular DL.'' By ``regular DL'' we mean that non-zero dGDVMs are used to initialize a DL architecture that is based on a ``standard'' neural network, but the architecture itself does not operate on the dynamic PSNs (in contrast to graph-based DL described later in Section~\ref{sect:GCN} that explicitly operates on the PSNs). Note that in the non-zero dGDVM of a PSN, nodes are ordered as they occur in the protein sequence; this is exactly why we can use regular DL (i.e. a ``standard'' neural network) on PSNs in the first place.

In more detail, the regular DL architecture that we consider combines CNNs and bidirectional LSTM layers. CNN layers apply learnable filters (kernels) over small neighborhoods of an input matrix to detect recurring patterns. In our setting, the input is an $n \times 211$ (i.e. non-zero) dGDVM. A CNN layer examines small groups of neighboring rows (nodes in the PSN) while spanning all columns (graphlet orbits), enabling the model to detect structural patterns across nearby nodes in the dynamic PSN over all graphlet orbits. Stacking multiple CNN layers allows the model to capture increasingly complex patterns by combining information from progressively larger neighborhoods of nodes. CNN layers can accommodate protein domains with different numbers of rows in non-zero dGDVMs because convolutional filters are applied locally and repeatedly across the input matrix, rather than being tied to one fixed input length; CNN layers are useful for identifying local combinations of graphlets that occur among nearby amino acids or nearby positions in the non-zero dGDVM. Following the CNN layers, we employ LSTM layers. Like CNN layers, LSTM layers can also be applied to dGDVMs with variable numbers of rows, since they process the input as a sequence whose length is determined by $n$. However, CNN and LSTM layers serve different purposes. CNN layers focus on detecting local patterns by applying filters over small neighborhoods of rows and feature columns. In contrast, LSTM layers model sequential dependencies across the ordered rows of the dGDVM. Specifically, an LSTM reads the row-level representations one step at a time and updates a hidden state as each row is processed, allowing information from earlier rows to influence the interpretation of later rows. Therefore, while CNN layers learn local graphlet-feature patterns, LSTM layers learn how these patterns or row-level representations are related across the longer ordered sequence of amino acids in the domain. The use of a \textit{bidirectional} LSTM allows the model to incorporate contextual information from both preceding and subsequent nodes in the sequence (i.e. rows in the dGDVM).

Our regular DL architecture takes as input an $n \times 211$ (non-zero) dGDVM. For simplicity, we refer to this input feature as ``dynamic graphlets.'' We denote the regular DL architecture as ``dynamic graphlets + ($x$ CNN, $y$ LSTM)'', where $x$ and $y$ indicate the numbers of CNN and LSTM layers, respectively. To assess how architectural design influences PSC performance, we design four variants of the regular DL architecture that vary several  parameters (unless explicitly noted otherwise, a variant uses the ReLU activation function): 
\begin{itemize}
    \item The first regular DL variant that we consider is dynamic graphlets + (2 CNN, 3 LSTM) under ReLU activation. We use this variant because our lab's previous collaborative work \cite{guo2019weighted} showed that a regular DL design having two CNN layers and three LSTM layers was effective for the task of PSC using \textit{static} PSNs.
    \item The second regular DL variant that we consider is dynamic graphlets + (3 CNN, 3 LSTM) under ReLU activation. To evaluate the effect of CNN depth, we compare the dynamic graphlets + (2 CNN, 3 LSTM) variant with the dynamic graphlets + (3 CNN, 3 LSTM) variant, which increases the number of CNN layers  under the same activation function (ReLU) (i.e. the two variants are the same except the number of CNN layers).
    \item The third regular DL variant that we consider is dynamic graphlets + (3 CNN, 1 LSTM) under ReLU activation. To evaluate the effect of recurrent depth, we compare the dynamic graphlets + (3 CNN, 3 LSTM) variant with the dynamic graphlets + (3 CNN, 1 LSTM) variant, which reduces the number of LSTM layers while keeping the CNN structure unchanged (i.e. the two variants are the same except the number of LSTM layers).
    \item The fourth regular DL variant that we consider is dynamic graphlets + (3 CNN, 1 LSTM, LeakyReLU). To evaluate the influence of nonlinear activation functions, we compare the dynamic graphlets + (3 CNN, 1 LSTM) variant, which uses ReLU activation as stated above,  with the dynamic graphlets + (3 CNN, 1 LSTM, LeakyReLU) variant, in which ReLU activation is replaced with LeakyReLU activation (i.e. the two variants are the same except the activation function). 
\end{itemize}

Clearly, the four considered variants are not an exhaustive list of all possible combinations of the considered parameters. While it would have been ideal to consider the exhaustive list, this would have been  extremely challenging due to high runtime of DL (as our results later on in the paper will show). Hence, we have aimed to do as comprehensive and systematic search over as small subset as possible of all combinations of the considered parameters that would still allow us to vary only a single parameter at a time when we evaluate the effect of the choice of the number of CNN layers, the number of LSTM layers, or the activation function. 

All four considered variants use the same $i$ non-zero dGDVM inputs ($i$ matrices for $i$ protein domains/dynamic PSNs in a given dataset), optimization strategy, parameter settings, hyperparameter tuning, and cross-validation framework to ensure that observed performance differences arise only from architectural changes. 

We illustrate our regular DL variants in Fig. \ref{fig:fig_methods}(c), and we explain them in more details in \ref{sect:sup_CNN_LSTM}.

\subsubsection{The graph-based DL architecture and its variants}\label{sect:GCN}

We next examine whether DL can do more than learn ``only'' from graphlet feature matrices (as regular DL does); that is, we examine whether DL can also learn directly from graph structures (another DL paradigm that we refer to as ``graph-based DL'' -- see below). In the regular DL framework described above, the input is the raw non-zero dGDVM extracted from a dynamic PSN; thus, learning is performed on precomputed dynamic graphlet features rather than directly on the PSN itself. In order to learn directly from a dynamic PSN, we employ graph-based DL. By ``graph-based DL,'' we mean DL architectures that take dynamic PSNs themselves as input and operate on their graph structure. These models may also use node-level features derived from the dynamic PSNs (i.e. non-zero dGDVMs), but the key distinction from regular DL is that graph-based DL learns directly from the PSN topology rather than only on precomputed feature matrices. A prominent method type within the graph-based DL paradigm are graph neural networks (GNNs). In this paper, we focus specifically on GNNs and, within this broader class, on graph convolutional networks (GCNs), which are among the most widely used GNN architectures for learning from graph-structured data. In the context of PSNs, GCNs learn representations by iteratively aggregating information across  neighboring nodes (i.e. amino acids). Through multiple layers of aggregation, more information can be collected from increasingly larger multi-hop neighborhoods. In this way, GCNs can learn structural features explicitly from the network connectivity itself.

In our study, we consider two particular GCN variants that operate on either dynamic or static PSNs (we justify why we consider the latter in more detail below). For the first variant that operates on dynamic PSNs: because dynamic PSNs consist of a temporal series of network snapshots (mimicking the process of protein folding dynamics), we employ in our study a discrete-time dynamic graph model. This model extends standard GCNs to discrete-time dynamic graphs by jointly modeling spatial relationships within individual snapshots and temporal dependencies across successive snapshots \cite{feng2025comprehensive}. More specifically, we use a dynamic GCN (DGCN) \cite{manessi2020dynamic}. It is of note that DGCNs require node-level features as input, in addition to the graph adjacency structure. To examine the influence of feature initialization, we consider two settings. In the first feature initialization setting, to allow the DGCN to learn representations based solely on the PSN topology without relying entirely on pre-engineered structural descriptors, node features are randomly initialized (referred to as \emph{default} features). In the second feature initialization setting, to allow the model to incorporate predefined higher-order topological descriptors together with the dynamic PSN structure during learning, node features are initialized using dynamic graphlets derived from the dynamic PSN (i.e. the non-zero dGDVM). In our study, we refer to the DGCN variant that utilizes default features as ``default features + DGCN'' and to the DGCN variant that utilizes dynamic graphlets as ``dynamic graphlets + DGCN.'' By comparing these two DGCN variants, we can assess whether graphlet-based features provide additional predictive signal beyond what can be learned directly from the network structure.

For the second variant that operates on static PSNs: because the temporal modeling component of the DGCN explicitly captures dependencies across multiple dynamic PSN snapshots, it introduces additional model complexity and parameters that may increase variance or amplify noise (as described in more detail in \ref{sect:sup_GCN}). As a result, such added complexity and parameters could instead degrade PSC performance. To assess whether modeling temporal graph evolution improves PSC performance, we also evaluate a static GCN (SGCN). The SGCN explicitly operates only on a single PSN snapshot corresponding to the final native protein structure while retaining all other properties as the DGCN. We refer to this SGCN variant as ``dynamic graphlets + SGCN.'' By comparing this graph-based DL variant to ``dynamic graphlets + DGCN,'' we can isolate and evaluate the effect of temporal modeling in dynamic PSNs via the DGCN. 

We illustrate our graph-based DL variants in Fig. \ref{fig:fig_methods}(d) and explain them in more detail in \ref{sect:sup_GCN}.

    \subsection{Evaluation}\label{sect:evaluation}

\subsubsection{Training and hyperparameter optimization}

We evaluate considered methods on the 72 datasets under our established protocol \cite{newaz2022multi} that allows for a fair comparison between traditional ML, regular DL, and graph-based DL. Namely, we ensure that potential PSC performance differences reflect methodological differences rather than inconsistencies in evaluation setup.

In more detail, for each dataset, we train and test all methods on identical training and test splits during classification via five-fold cross-validation. Every dataset is partitioned into five approximately equal-sized folds, with the proportion of structural classes preserved within each fold to maintain class balance. For each of the five iterations, one fold is used as the test set and the remaining four folds are combined to form the training set (in other words, every protein domain in a dataset is used exactly once for testing and four times for training).

Moreover, we employ nested five-fold cross-validation \emph{during training} for hyperparameter tuning, by partitioning the training set into five new (inner) folds using the same partitioning strategy as above. For each candidate hyperparameter value, the predictive model is trained on four inner folds and validated on the remaining inner fold, rotating through all five combinations. The value yielding the lowest misclassification rate (see below) during inner validation  is selected as the optimal hyperparameter.

\subsubsection{Evaluation measure -- misclassification rate}

For each dataset, for each method, we asses classification performance using misclassification rate. For a given set of predictions, misclassification rate is the portion of the predictions (in our case, protein domains, i.e. dynamic PSNs) whose predicted classes do \emph{not} match their ground-truth labels (in our case, CATH/SCOPe structural classes). The lower  the misclassification rate, the better. Note that we express misclassification rate in the [$0,1$] range (rather than in the [$0\%,100\%$] range), and we round it to the third decimal place.

One could compute misclassification rate on the test fold in each iteration of the five-fold cross-validation and compute their average. Or, one could compute the aggregate misclassification rate over all five test folds combined. The latter is simpler to visualize and interpret \cite{guo2019weighted}. More importantly, in our prior work on the same PSC task using the same data and evaluation framework, we observed no differences in results between the average and aggregate misclassification rates. We have verified the same in this current study as well (as we briefly discuss in Section \ref{sect:results}). This is why when we report results, we adopt the aggregate misclassifications rate as our main evaluation measure  \cite{newaz2022multi}. 

\subsubsection{Method ranking}

We summarize method performance over all 72 datasets as follows. For a given group of methods being compared, for a given dataset, we rank the methods in that group relative to one another from lowest (best, rank 1) to highest (worst) misclassification rate. In more detail, we use a competition-style ranking scheme in which lower misclassification corresponds to better rank and ties receive the same rank, but subsequent ranks reflect the total number of methods that precede them (i.e. ranks may be skipped). For example, if three methods are compared on a given dataset and two achieve the same lowest misclassification rate, the two methods are all assigned rank 1, and the remaining method is assigned rank 3. For a given group of methods being compared, after we compute their per-dataset rankings as described, we then summarize each method’s performance across all 72 datasets. Specifically, for a given method, we calculate the percentage of the 72 datasets on which it is ranked as the best (rank 1); in the process, we report how many of those best ranks are ``absolute rank 1'' (without ties to any other method in the comparison group on a given dataset) vs. ``tied rank 1''  (a tie with at least one other method in the comparison group on a given dataset). 

To avoid claiming any method to be superior over another method due to only a negligible difference in the methods' misclassification rates, we use two different ``definitions'' of a tie, i.e. two ranking ``policies'', referred to as strict ranking and relaxed ranking. Under strict ranking, two methods are considered to be tied if the absolute difference between their misclassification rates is exactly 0 (i.e. 0\%). Under relaxed ranking, two methods are  considered to be  tied if the absolute difference between their misclassification rates is less than or equal to 0.02 (i.e. 2\%). For example, 
if a method achieves  misclassification rate of 0.070, and another method achieves a misclassification rate of 0.088, the absolute difference between their classification rates is $0.088 - 0.070 = 0.018$, and so under relaxed ranking, the two methods would be considered tied. Note that in addition to 2\%, we tested relaxed ranking for 1\%, 5\%, and 10\% as well, and the general trends in results are the same as for 2\% (results not shown). 

\subsubsection{Statistical significance of method performance}

Beyond the ranking analysis, we also evaluate whether differences in misclassification rates across the 72 datasets between pairs of methods are statistically significant. Specifically, for each pair of methods, we construct 72 paired observations consisting of their (aggregate) misclassification rates for the 72 datasets, and we perform a paired statistical test -- one-sided paired Wilcoxon sign-rank -- across the paired observations. Because we conduct multiple pairwise comparisons -- corresponding to all pairs of considered methods (across all three traditional ML and DL paradigms) -- we apply a Bonferroni correction to control the family-wise error rate. We report corrected $p$-values, i.e. $q$-values.

\subsubsection{Runtimes}

For a given method, in addition to PSC classification performance (i.e. misclassification rate), we also evaluate its computational runtime. Specifically, we report a given method's classification runtime on each individual dataset, and we also report its summarized classification runtime (median, mean, and standard deviation) across all 72 datasets. For the regular DL variants, all analyses are run on 25 AMD Opteron Processor 6376 cores. For the LR architecture and the graph-based DL variants, all analyses are executed on 25 AMD EPYC 7543 32-Core Processor cores.

\section{Results}\label{sect:results}

\subsection{Different variants of regular DL}\label{sect:results_compare_DL}

To assess the effect of architectural design choices within the regular DL paradigm, we compare its four CNN+LSTM variants, which differ in the number of CNN/LSTM layers and activation function, but not initialization features (Table \ref{tab:list_of_methods}). 

\subsubsection{Effect of different activation functions (under the same number of CNN and LSTM layers)}

To evaluate the impact of nonlinear transformation (via activation functions) independently of model depth, we compare the two architectural variants with identical numbers of CNN and LSTM layers (three and one, respectively) but different activation functions (ReLU vs. LeakyReLU). 

\begin{figure*}[ht!]
    \begin{center}
            \includegraphics[
                width=1\textwidth,
                trim= 1cm 11cm 1cm 0cm,
                clip
            ]{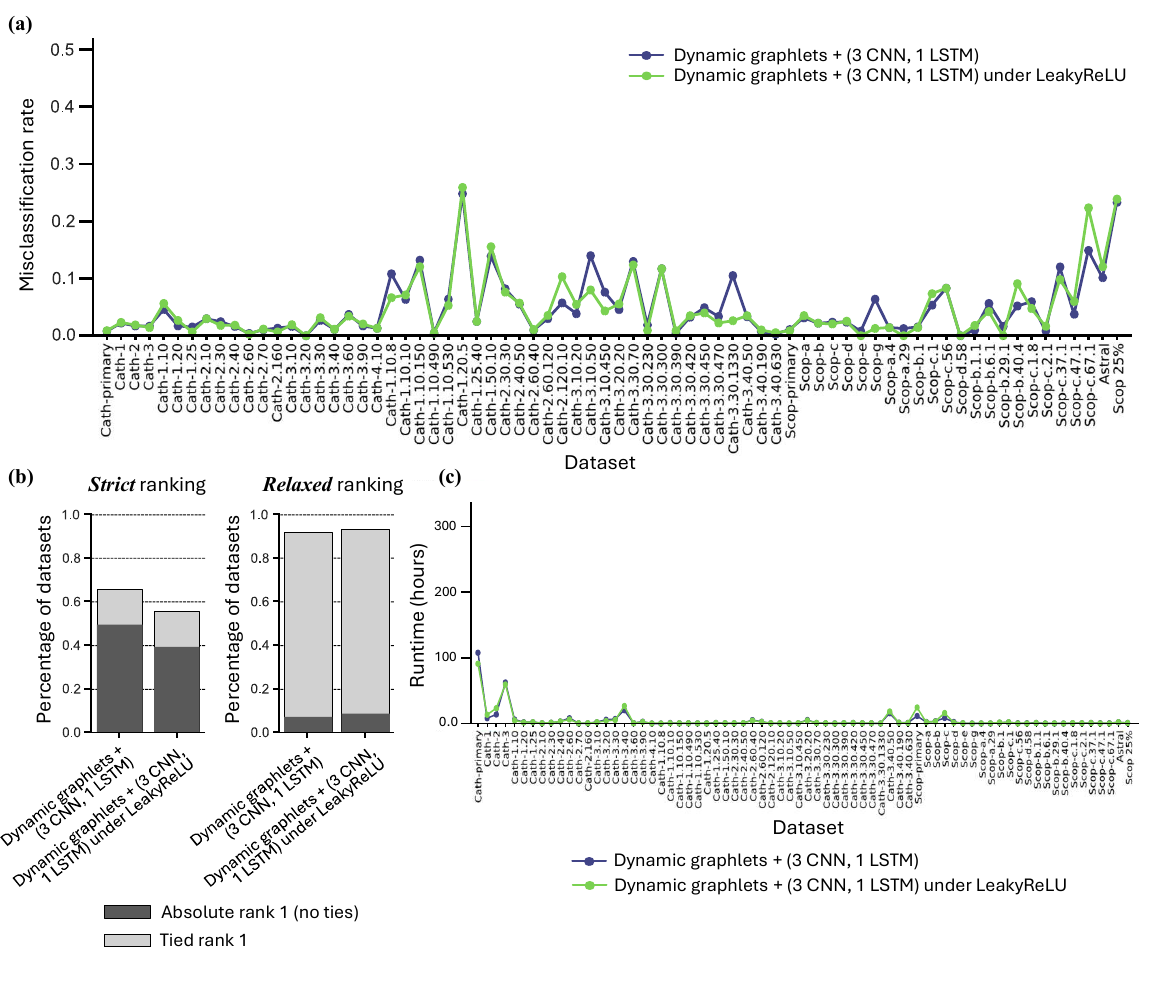}%
    \end{center}
    \caption{Comparison of the ReLU and LeakyReLU activation functions under the same number of CNN and LSTM layers within the regular DL paradigm on the 72 datasets in the task of dynamic PSN-based PSC. Specifically, the   
    dynamic graphlets + (3 CNN, 1 LSTM) variant under ReLU and the dynamic graphlets + (3 CNN, 1 LSTM) variant under LeakyReLU are compared. \textbf{(a)} Per-dataset misclassification rates; recall that we report rates aggregated over all test folds of the 5-fold cross-validation. \textbf{(b)} Left: The percentage of all 72 datasets in which a given method is the best (rank 1), or is tied as the best (within 0\% absolute difference aka strict ranking). Right: The percentage of all 72 datasets in which a given method is the best (rank 1), or is close-to-tied as the best (within 2\% absolute difference aka relaxed ranking). \textbf{(c)} Per-dataset runtime (in hours).}
    \label{fig:sup_fig_1}
\end{figure*}

In terms of misclassification rate, the ReLU and leakyReLU variants perform quite comparably for many of the 72 datasets (Supplementary Fig. \ref{fig:sup_fig_1}(a)). When for more clarity we examine the percentage of the 72 datasets on which a given method is the best or is \emph{strictly} tied as the best (\emph{strict} ranking; Section \ref{sect:evaluation}), ReLU is somewhat superior (Supplementary Fig. \ref{fig:sup_fig_1}(b)). When we examine the percentage of the 72 datasets on which a given method is the best or is \emph{close-to-tied} as the best (within 2\% absolute difference; \emph{relaxed} ranking; Section \ref{sect:evaluation}), ReLU is better than leakyReLU or is (much more often) close-to-tied with leakyReLU on 92\%
of the datasets. This result for leakyReLU over ReLU is 93\%. 
Hence, there exist lots of rank ties between the two  (Fig. \ref{fig:results_main}(a)). Note that relaxed ranking may be a better way to compare methods without claiming any one method to be superior due to only a negligible (within 2\%) absolute difference in misclassification rates.
When we evaluate whether a method is statistically outperforming another method over all 72 datasets, we find no significant difference between ReLU and leakyReLU (adjusted $p$-value, i.e. $q$-value, of 1; Fig. \ref{fig:results_main}(b,c)). In terms of runtimes, the two variants are comparable  (Fig. \ref{fig:results_main}(d) and  Supplementary Fig. \ref{fig:sup_fig_1}(c)). 

We conclude that the choice of activation function has no effect on PSC performance. Because ReLU is as good of a choice as leakyReLU, henceforth, we proceed with ReLU. 

\subsubsection{Effect of differing numbers of CNN and LSTM layers (under the same activation function)} 

To examine the influence of model depth, we compare the three architectures that vary in the number of CNN and LSTM layers -- (2 CNN, 3 LSTM), (3 CNN, 3 LSTM), (3 CNN, 1 LSTM) -- but that share the activation function (ReLU, per the above discussion); Table \ref{tab:list_of_methods}. 

\begin{figure*}[ht]
    \begin{center}
            \includegraphics[
                width=1\textwidth,
                trim= 0.5cm 11cm 1cm 0cm,
                clip
            ]{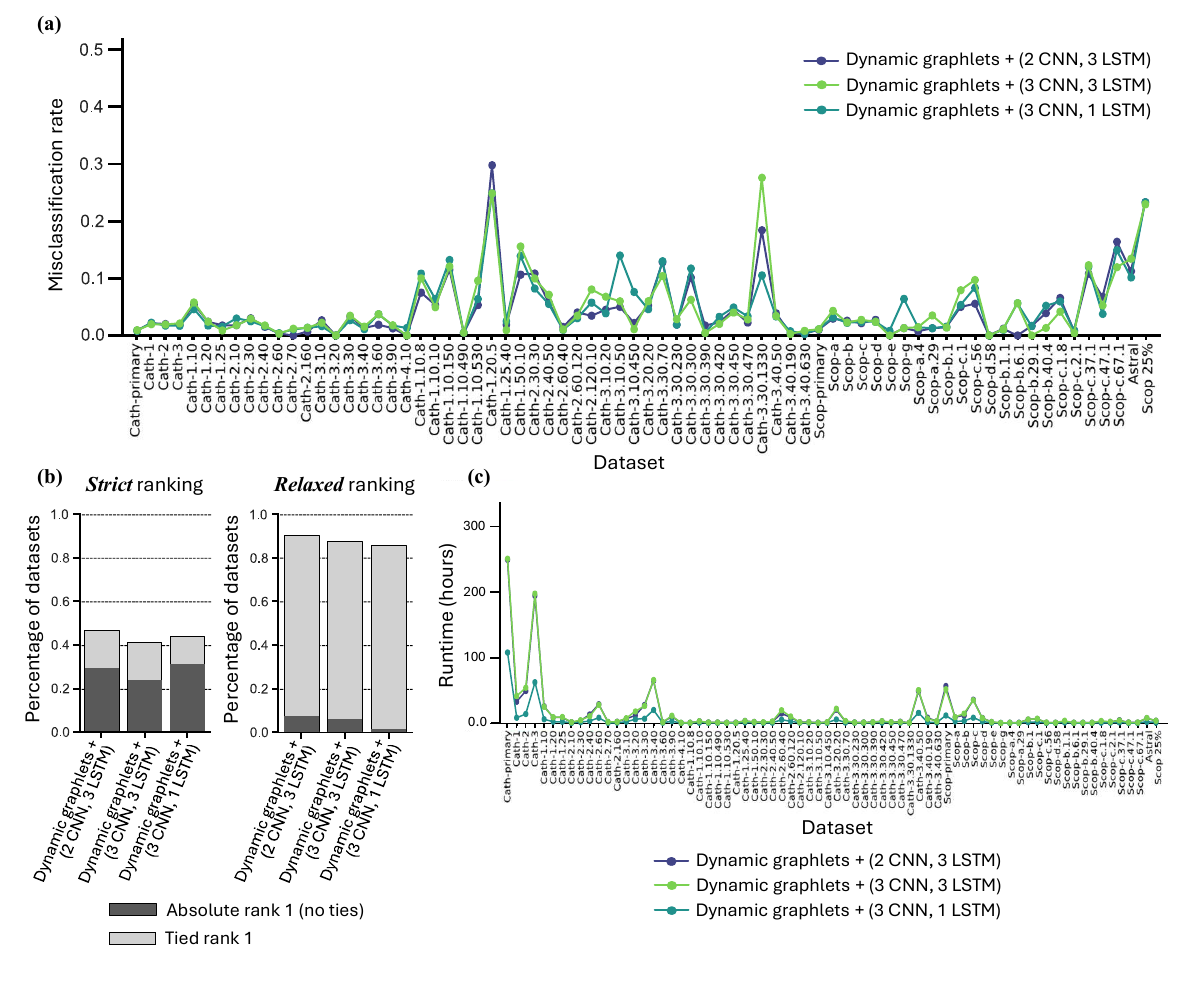}%
    \end{center}
    \caption{Comparison of the different numbers of CNN or LSTM layers under the same activation function (ReLU) within the regular DL paradigm on the 72 datasets in the task of dynamic PSN-based PSC. Specifically, the dynamic graphlets + (2 CNN, 3 LSTM) variant, the dynamic graphlets + (3 CNN, 3 LSTM), and the dynamic graphlets + (3 CNN, 1 LSTM) variant are compared. \textbf{(a)} Per-dataset misclassification rates; recall that we report rates aggregated over all test folds of the 5-fold cross-validation. \textbf{(b)} Left: The percentage of all 72 datasets in which a given method is the best (rank 1), or is tied as the best (within 0\% absolute difference aka strict ranking). Right: The percentage of all 72 datasets in which a given method is the best (rank 1), or is close-to-tied as the best (within 2\% absolute difference aka relaxed ranking). \textbf{(c)} Per-dataset runtime (in hours).}
    \label{fig:sup_fig_2}
\end{figure*}

In terms of misclassification rate, the three variants perform quite comparably for many but not all datasets (Supplementary Fig. \ref{fig:sup_fig_2}(a)).  For further clarity, we compare the variants' rankings over the 72 datasets. Under \emph{strict} ranking (already described), (2 CNN, 3 LSTM) is somewhat superior but all three variants have value on certain datasets (Supplementary Fig. \ref{fig:sup_fig_2}(b)). Under \emph{relaxed} ranking (already described), (2 CNN, 3 LSTM) is again somewhat superior (Fig. \ref{fig:results_main}(a)). In fact, (2 CNN, 3 LSTM) is better than,  or is (much more often) close-to-tied with, the other two variants on 89\% of the datasets, even though the other two variants closely follow (Fig. \ref{fig:results_main}(a)). None of the three  variants significantly outperforms over all 72 datasets any of the other two variants  (all pairwise q-values of 1; Fig. \ref{fig:results_main}(b,c)). 

We conclude that (2 CNN, 3 LSTM) is at least as good of a choice as (and actually at least somewhat superior to) the other two variants. Thus, we proceed with (2 CNN, 3 LSTM) as the overall best variant  within the regular DL paradigm, even though its runtime is not the fastest (Fig. \ref{fig:results_main}(d) and  Supplementary Fig. \ref{fig:sup_fig_2}(c)). 

In more detail regarding  runtimes, (3 CNN, 1 LSTM) is the fastest, followed by (2 CNN, 3 LSTM) and (3 CNN, 3 LSTM) that are comparable to one another. The fact that (2 CNN, 3 LSTM) and (3 CNN, 3 LSTM) have comparable runtimes indicates that the number of CNN layers has no effect. The fact that (3 CNN, 1 LSTM) and (3 CNN, 3 LSTM) have quite differing runtimes indicates that the number of LSTM layers does have an effect.
    
\subsection{Different variants of graph-based DL}\label{sect:results_compare_GCN}

Next, we evaluate the three graph-based DL variants 
(Table \ref{tab:list_of_methods}). We assess two separate effects.

\subsubsection{Effect of different initialization features (under the same GCN)} We examine the effect of initializing a GCN with its default vs.  dynamic graphlet features. That is, we compare default features + DGCN  against dynamic graphlets + DGCN. 

\begin{figure*}[ht]
        \begin{center}
        \includegraphics[width=0.96\textwidth,trim= 0.9cm 11.6cm 0.9cm 0cm]{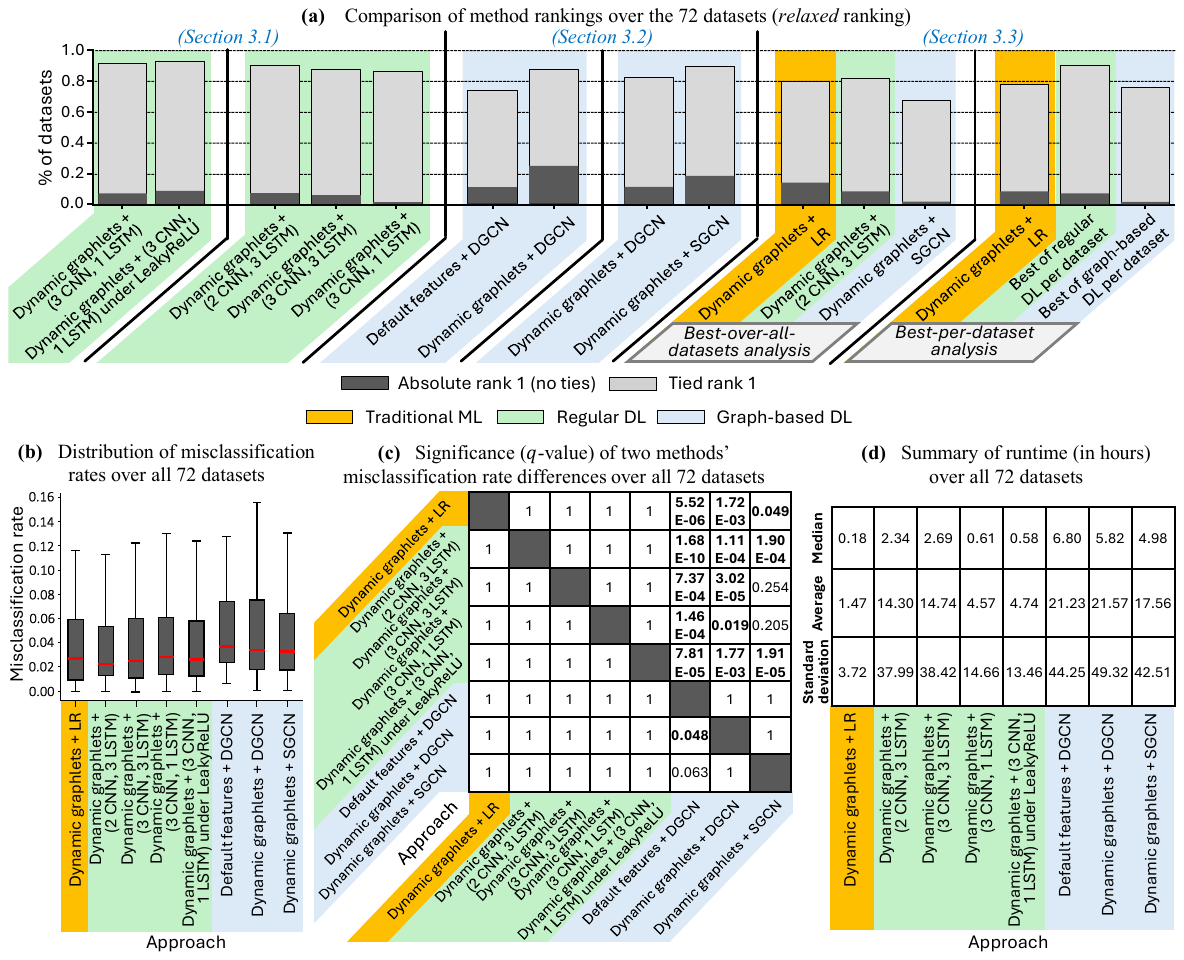}
        \end{center}
        \vspace{-0.4cm}
        \caption{Results of our comprehensive PSC evaluation summarized over the 72 datasets. Due to space constraints, for detailed, per-dataset results, see Supplementary Figs. \ref{fig:sup_fig_1}-\ref{fig:sup_fig_6}. \textbf{(a)} The percentage of all 72 datasets in which a given method is the best (rank 1), or is close-to-tied as the best (within 2\% absolute difference aka \emph{relaxed} ranking). The six  parts correspond to the six analyses, i.e. six groups of methods being evaluated, in the indicated paper sections. Only within a given analysis/method group, the methods from that group are compared to each other. For any (CNN, LSTM) approach variant, unless the LeakyReLU activation function is explicitly listed in the variant's name, it is the ReLU activation function that is used. \textbf{(b)} The distribution of a method's misclassification rates over all 72 datasets. The red line is the distribution mean. For clarity, we left out outliers from this visualization; for the results with the outliers, see Supplementary Fig. \ref{fig:sup_fig_7}. \textbf{(c)} Cell $(i,j)$ contains the $q$-value indicating whether method $i$ yields significantly better (lower) misclassification rates than method $j$ when accounting for the methods' PSC performance over all 72 datasets; $q$-values $< 0.05$ are bolded. \textbf{(d)} A method's median and average runtime over all 72 datasets. The standard deviations are large because the different datasets can vary greatly in size and thus in runtime.}
        \label{fig:results_main}
    \end{figure*}

\begin{figure*}[ht]
    \begin{center}
            \includegraphics[
                width=1\textwidth,
                trim= 1cm 11cm 1cm 0cm,
                clip
            ]{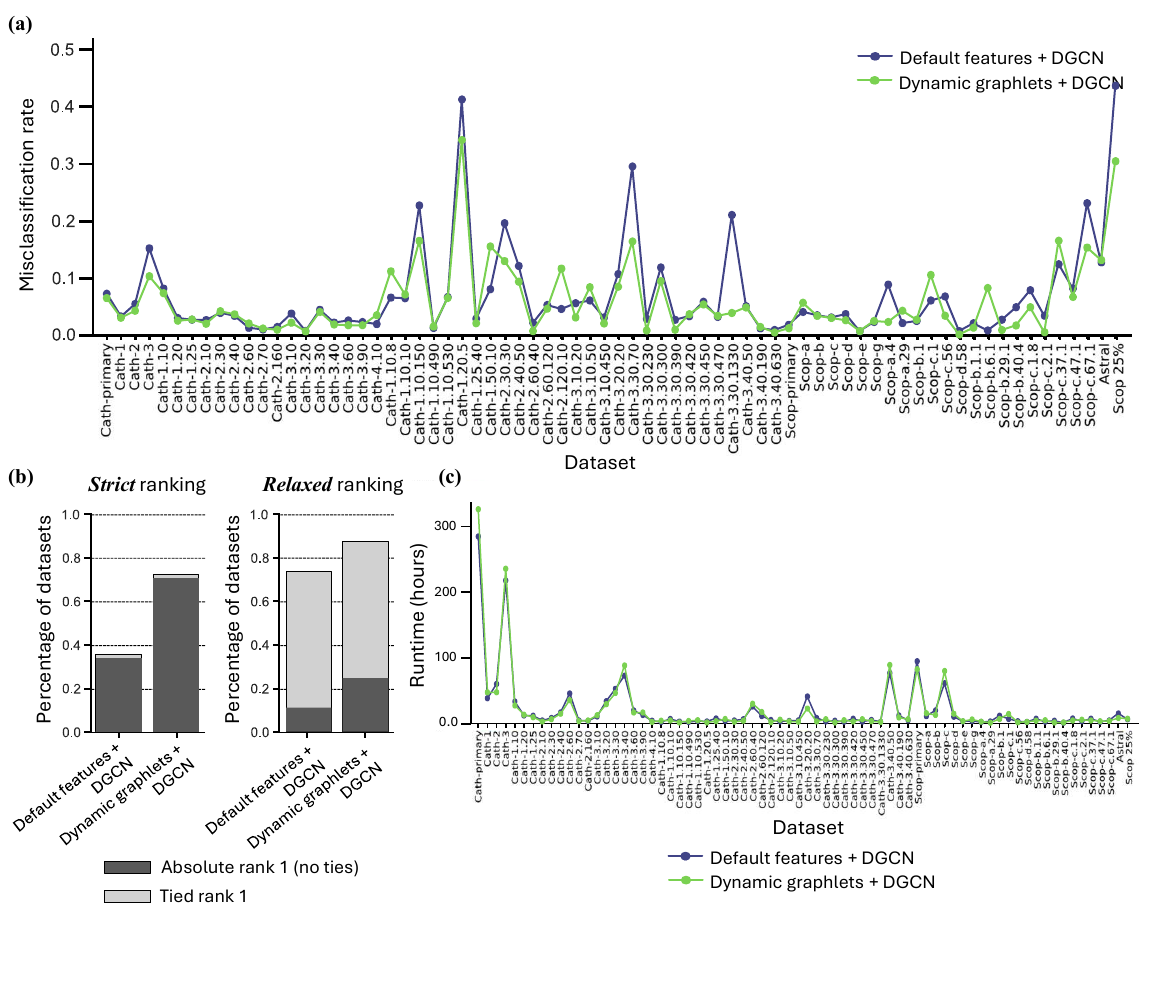}%
    \end{center}
    \caption{Comparison of default features and dynamic graphlets under the same GCN (DGCN) within the graph-based DL paradigm on the 72 datasets in the task of dynamic PSN-based PSC. Specifically,   
    the default features + DGCN variant and the dynamic graphlets + DGCN variant are compared. \textbf{(a)} Per-dataset misclassification rates; recall that we report rates aggregated over all test folds of the 5-fold cross-validation. \textbf{(b)} Left: The percentage of all 72 datasets in which a given method is the best (rank 1), or is tied as the best (within 0\% absolute difference aka strict ranking). Right: The percentage of all 72 datasets in which a given method is the best (rank 1), or is close-to-tied as the best (within 2\% absolute difference aka relaxed ranking). \textbf{(c)} Per-dataset runtime (in hours).}
    \label{fig:sup_fig_3}
\end{figure*}

In terms of misclassification rate, the per-dataset  results are not obvious (Supplementary Fig. \ref{fig:sup_fig_3}(a)). For clarity, we examine the variants' rankings over the 72 datasets. Under \emph{strict} ranking, dynamic graphlets show a large superiority over default features (Supplementary Fig. \ref{fig:sup_fig_3}(b)). Under \emph{relaxed} ranking, dynamic graphlets remain superior: they are  better than default features or are (more often) close-to-tied with default features on 88\% of the datasets; this result for default features over dynamic graphlets is 73\% (Fig. \ref{fig:results_main}(a)). 
The superiority of dynamic graphlets over default features across all 72 datasets is significant ($q$-value of $7.3 \times 10^{-4}$; Fig. \ref{fig:results_main}(b,c)). 

As dynamic graphlets are better GCN initialization features, we proceed with them. This is further justified by the fact that the two feature types have comparable runtimes (Fig. \ref{fig:results_main}(d) and  Supplementary Fig. \ref{fig:sup_fig_3}(c)).

\subsubsection{Effect of different GCNs (under the same initialization feature)}

We analyze \emph{dynamic} PSNs. Thus, it seems logical to use DGCN (i.e. \emph{dynamic} GCN). Yet, recall that to use DGCN on dynamic PSNs whose temporal snapshots are of different sizes, we have to use padding, which is a heuristic (\ref{sect:sup_GCN}). Hence, while DGCN should capture network dynamics better than SGCN (i.e. \emph{static} GCN), this is not guaranteed. Thus, we compare the DGCN variant and the SGCN variant (both initialized with dynamic graphlet features, per the above discussion), to assess whether DGCN actually provides gain compared to SGCN (as we design and implement them). Importantly, even though SGCN is static, because it is initialized by dynamic graphlet features, SGCN is still capturing the dynamics of the PSNs. 

\begin{figure*}[ht]
    \begin{center}
            \includegraphics[
                width=1\textwidth,
                trim= 1cm 11cm 1cm 0cm,
                clip
            ]{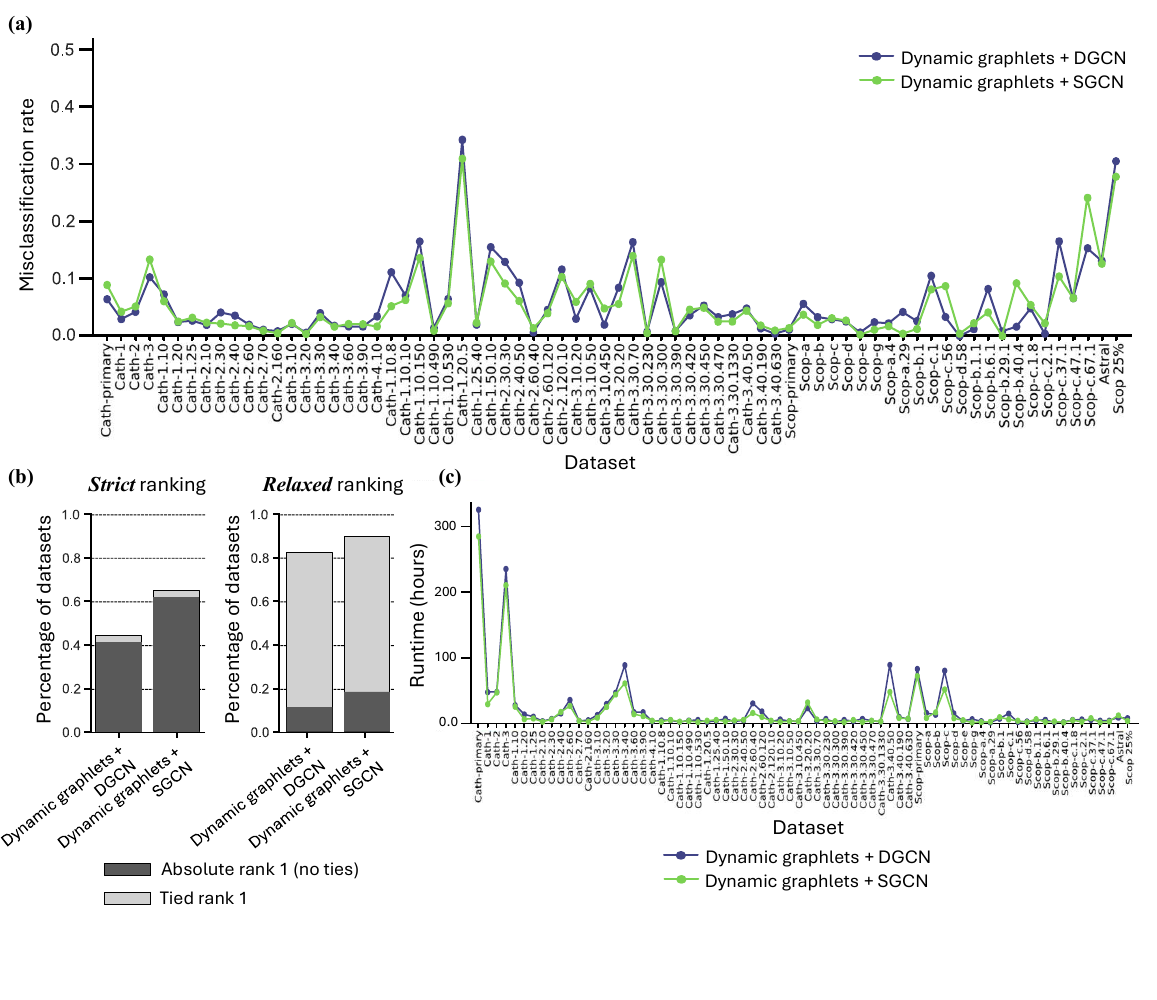}%
    \end{center}
    \caption{Comparison of DGCN and SGCN under the same features (dynamic graphlets) within the graph-based DL paradigm.    
    Specifically, the dynamic graphlets + DGCN variant and the dynamic graphlets + SGCN variant are compared. \textbf{(a)} Per-dataset misclassification rates; recall that we report rates aggregated over all test folds of the 5-fold cross-validation. \textbf{(b)} Left: The percentage of all 72 datasets in which a given method is the best (rank 1), or is tied as the best (within 0\% absolute difference aka strict ranking). Right: The percentage of all 72 datasets in which a given method is the best (rank 1), or is close-to-tied as the best (within 2\% absolute difference aka relaxed ranking). \textbf{(c)} Per-dataset runtime (in hours).}
    \label{fig:sup_fig_4}
\end{figure*}

In terms of misclassification rate, the per-dataset results are not obvious (Supplementary Fig. \ref{fig:sup_fig_4}(a)). Hence, for clarity, we examine the variants' rankings over the 72 datasets. Under \emph{strict} ranking, SGCN shows  superiority over DGCN, but both have value on certain datasets (Supplementary Fig. \ref{fig:sup_fig_4}(b)). Under \emph{relaxed} ranking, SGCN is still somewhat superior: it is  better than DGCN or is (much more often) close-to-tied with DGCN on 89\% of the datasets (Fig. \ref{fig:results_main}(a)). This result for DGCN over SGCN is 82\%. Hence, there exist lots of rank ties between the two (Fig. \ref{fig:results_main}(a)). None of the two variants significantly  outperforms the other ($q$-value of 1; Fig. \ref{fig:results_main}(b,c)). 

Given somewhat better performance of SGCN in terms of misclassification rate and somewhat higher runtimes of DGCN  (Fig. \ref{fig:results_main}(d) and Supplementary Fig. \ref{fig:sup_fig_4}(c)), we proceed with  dynamic graphlets + SGCN as the overall best  graph-based DL variant. 
    
\subsection{Traditional ML vs. regular DL vs. graph-based DL}\label{sect:results_compare_overall}

We next compare the (i) traditional ML approach (dynamic graphlets + LR), (ii) overall best regular DL variant (dynamic graphlets + (2 CNN, 3 LSTM) under ReLU), and (iii) overall best graph-based DL variant (dynamic graphlets + SGCN). This is motivated by  \underline{our key question}: whether increasing model complexity -- from manual, pre-engineered features under an off-the-shelf classifier, to deep neural networks initialized by the same features, to end-to-end graph representation learning -- yields improvement in the PSC task. 

\begin{figure*}
        \begin{center}
        \includegraphics[width=1\textwidth, trim= 0.8cm 20.8cm 0.8cm 0cm]{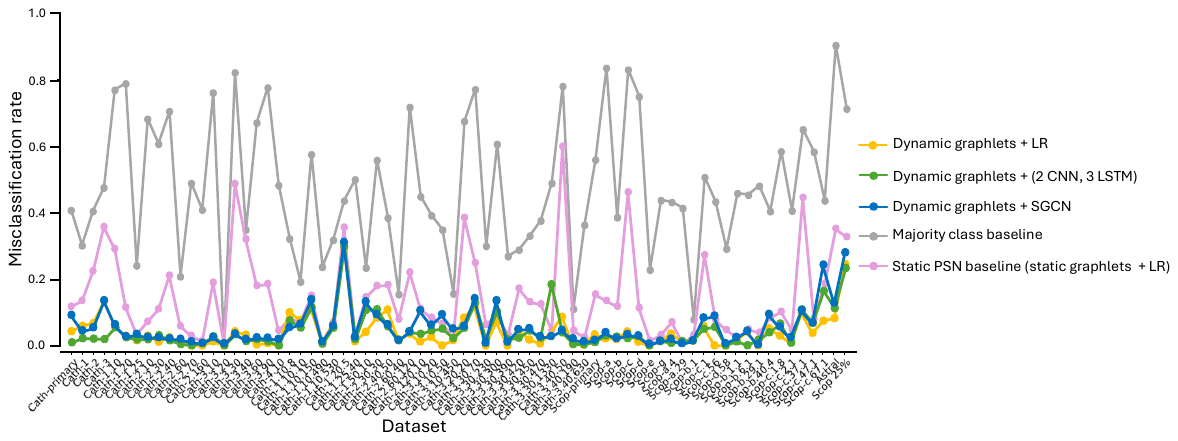}
        \end{center}
        \caption{Per-dataset misclassification rates in the best-over-all-datasets analysis; recall that we report rates aggregated over all test folds of the 5-fold cross-validation. The key comparison is between traditional ML (dynamic graphlets + LR), the overall best variant in the regular DL paradigm (dynamic graphlets + (2 CNN, 3 LSTM)), and the overall best variant in the graph-based DL paradigm (dynamic graphlets + SGCN). In this figure (and in its best-per-dataset analysis counterpart; Supplementary Fig. \ref{fig:sup_fig_6_alt}), to illustrate the power of these three PSC approaches that are all based on dynamic PSNs, we add results for two baselines that do not use dynamic PSNs. One is majority class, which is what the misclassification rate would be if all protein domains in a given dataset were predicted  to have the largest of all classes in the dataset. The other one is based on static PSNs \cite{newaz2020network,newaz2022multi}; this is the static counterpart of dynamic graphlets + LR, whose variant was already shown to be competitive to multiple sequence and direct (non-network) 3D structural approaches in the same PSC task  (Section \ref{sect:introduction} and Fig. \ref{fig:overview}(a)) \cite{newaz2020network}. Note that we do consider approaches for the task of protein structure comparison, e.g. TM-align \cite{Zhang2005TMalign} or Foldseek \cite{vanKempen2024Foldseek}, in our PSC task/evaluation, because the two tasks (their comparison vs. our classification) are distinct computationally and evaluation-wise, as discussed in the text.}
      \vspace{-0.3cm}
        \label{fig:results_over_all}
    \end{figure*}

\begin{figure*}[ht]
    \begin{center}
            \includegraphics[
                width=1\textwidth,
                trim= 0.5cm 11cm 1cm 0cm,
                clip
            ]{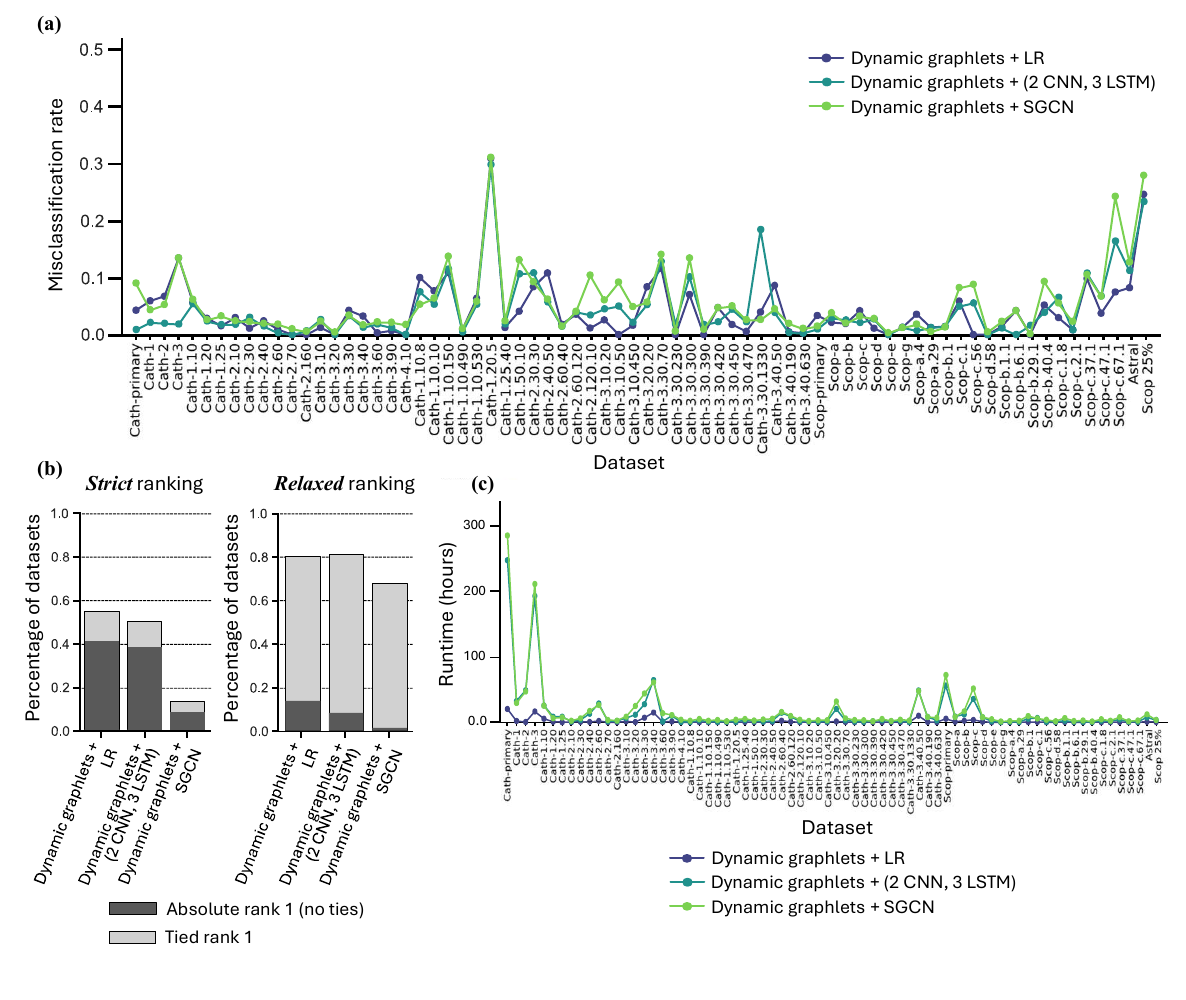}%
    \end{center}
    \caption{Comparison of traditional ML, regular DL, and graph-based DL in the \emph{best-over-all-datasets} analysis on 72 datasets in the task of dynamic PSN-based PSC. Specifically,   the 
    dynamic graphlets + LR approach, the dynamic graphlets + (2 CNN, 3 LSTM) variant, and the dynamic graphlets + SGCN variant are compared. \textbf{(a)} Per-dataset misclassification rates; recall that we report rates aggregated over all test folds of the 5-fold cross-validation. \textbf{(b)} Left: The percentage of all 72 datasets in which a given method is the best (rank 1), or is tied as the best (within 0\% absolute difference aka strict ranking). Right: The percentage of all 72 datasets in which a given method is the best (rank 1), or is close-to-tied as the best (within 2\% absolute difference aka relaxed ranking). \textbf{(c)} Per-dataset runtime (in hours).}
    \label{fig:sup_fig_5}
\end{figure*}

When we compare the three methods' misclassification rates,
the per-dataset results are not obvious (Fig. \ref{fig:results_over_all} and Supplementary Fig. \ref{fig:sup_fig_5}(a)). Thus, we examine their rankings over the 72 datasets. Under \emph{strict} ranking, traditional ML is superior to  both DL approaches (Supplementary Fig. \ref{fig:sup_fig_5}(b)). Under \emph{relaxed} ranking,  traditional ML and regular DL show a closer performance (Fig. \ref{fig:results_main}(a)). That is, dynamic graphlets + LR is  better than, or is (much more often) close-to-tied with, the other two approaches on 81\% of the datasets. This result for dynamic graphlets + (2 CNN, 3 LSTM) is 82\%, and for dynamic graphlets + SGCN it is 69\%  (Fig. \ref{fig:results_main}(a)). Hence, there exist lots of rank ties  (Fig. \ref{fig:results_main}(a)). Dynamic graphlets + SGCN is significantly outperformed by both dynamic graphlets + LR  ($q$-value of $0.049$) and dynamic graphlets + (2 CNN, 3 LSTM) ($q$-value of $1.9 \times 10^{-4}$); Fig. \ref{fig:results_main}(b,c). None of dynamic graphlets + LR and dynamic graphlets + (2 CNN, 3 LSTM) significantly outperforms the other ($q$-value of 1).

When we compare the methods' runtimes (Fig. \ref{fig:results_main}(d) and Supplementary Fig. \ref{fig:sup_fig_5}(c)), dynamic graphlets + LR, i.e.  traditional ML, is substantially faster than both regular and graph-based DL. Compared to dynamic graphlets + LR, on average over all datasets, dynamic graphlets + (2 CNN, 3 LSTM) incurs nearly 10 times greater runtime, and dynamic graphlets + SGCN incurs nearly 12 times greater runtime.

These findings demonstrate that in the task of dynamic PSN-based PSC, given the specific approaches and their parameters as well as evaluation that we have considered, both traditional ML and regular DL outperform graph-based DL, in terms of both misclassification rate and runtime. However, we find that regular DL does not guarantee gain over traditional ML. In fact, traditional ML holds its value, i.e. is closely comparable overall to regular DL in terms of misclassification rate, while also being much faster. 

Above, we have chosen, within each DL paradigm, the variant that is (close-to-tied as) the best on more datasets than the other variant(s). But, the chosen variant might not be the best for each dataset. For example, within the graph-based DL paradigm, dynamic graphlets + SGCN was chosen as the  overall best variant, but this is not the best graph-based DL variant for each of the 72 datasets. Hence, next, we compare traditional ML vs. regular DL vs. graph-based DL while considering, within each DL paradigm, the best variant \emph{per dataset} (as opposed to the best variant \emph{over all datasets}, as above).
For example, for regular DL, for all datasets where dynamic graphlets + SGCN is the best, we choose that variant, while for the other datasets where dynamic graphlets + DGCN is the best, we choose that variant.  

In this best-per-dataset setting, in terms of misclassification rates (Fig. \ref{fig:results_main}(a) and Supplementary Fig. \ref{fig:sup_fig_6_alt}), regular DL gains some superiority over traditional ML; yet the latter still holds its value. For example, under \emph{relaxed} ranking, dynamic graphlets + LR is still better than, or is (much more often) close-to-tied with, the two DL approaches on 79\% of the datasets; this result for regular DL is 90\%, 
and  for graph-based DL it is 76\% (Fig. \ref{fig:results_main}(a)). The now somewhat better performance of regular DL over traditional ML (yet with lots of ties) comes at the expense of an ever higher runtime of regular DL than in the best-over-all-datasets analysis. This is because in the latter,  one variant within a given DL paradigm is chosen over all datasets, and so we can consider only its runtime. On the other hand, in the best-per-dataset analysis, as different variants within a given DL paradigm are chosen for different datasets, the considered runtime for a given paradigm needs to account for combined runtimes over all of its variants. 

Regardless of whether looking at the best-over-all-datasets or best-per-dataset analysis, our findings demonstrate that no single learning paradigm dominates across all datasets with respect to misclassification rate in the task of dynamic PSN-based PSC. Regular DL and  traditional ML are relatively competitive to each other in terms of misclassification rate. Graph-based DL, while less frequently the best, still performs within a narrow margin of the other two categories. However, the computational trade-offs differ substantially across paradigms -- traditional ML is much more time-efficient across all datasets than both DL paradigms.  Of course, we again note that our findings and conclusions hold for the specific approaches and their parameters as well as evaluation that we have considered.

Our discussion thus far has focused on how the different paradigms compare in the PSC task when learning from \emph{dynamic PSNs}. While no single paradigm dominates across all datasets, all three paradigms when they use dynamic PSNs absolutely dominate baselines that are \emph{not} based on dynamic PSNs (Fig. \ref{fig:results_over_all} and Supplementary Fig. \ref{fig:sup_fig_6_alt}). This confirms the power of dynamic PSNs and dynamic graphlet features in the PSC task, regardless of the paradigm (traditional ML or DL). 

Note that one of the baselines that we consider is majority class -- what the misclassification rate would be if all protein domains in a given dataset were predicted  to have the largest of all classes in the dataset -- which can thus be seen as the very basic benchmark method to beat. The other considered baseline is based on static PSNs \cite{newaz2020network,newaz2022multi}; this is the static counterpart of dynamic graphlets + LR. We already showed that dynamic PSNs (dynamic graphlets + LR) are significantly superior to static PSNs (static graphlets + LR) in the PSC task \cite{newaz2022multi}, as already discussed in Section \ref{sect:introduction} and illustrated in Fig. \ref{fig:overview}(a). Recall also from that discussion/illustration that static PSNs were already shown to be superior or competitive to many sequence and direct (non-network) 3D structural approaches in our considered PSC task as well as in the related yet distinct tasks of protein structure comparison and protein function prediction. Specifically, in the case of the protein structure comparison task, static PSNs outperformed TM-align (Fig. \ref{fig:overview}(a)) \cite{faisal2017grafene}. While a new prominent approach for protein structure comparison has appeared more recently, namely Foldseek \cite{vanKempen2024Foldseek}, and it would thus be desireable to know how Foldseek compares to our dynamic PSN approach(es), their direct comparison is not appropriate as is. This is because  Foldseek is a protein structure \emph{comparison} approach  that compares \emph{pairs} of 3D structures and then on SCOPe data evaluates whether for the highest-similarity pairs, the domains in a given pair are members of the same SCOPe family \cite{vanKempen2024Foldseek}. On the other hand, the focus of our paper and thus of our use of dynamic PSN-based approach(es) is on the task of PSC -- using supervised \emph{classification} to predict the CATH/SCOPe label of an \emph{individual} domain based on its dynamic PSN feature. The two tasks (comparison vs. classification) are not fully independent but they have different purposes and are thus not directly comparable to each other; in fact,  similarities resulting from a protein structure comparison approach such as Foldseek can be used to cluster proteins in order to design  larger-scale benchmark data for our considered PSC task  \cite{Chen2026protein}.

\begin{figure*}[ht]
    \begin{center}
            \includegraphics[
                width=1\textwidth,
                trim= 1cm 10.5cm 1cm 0cm,
                clip
            ]{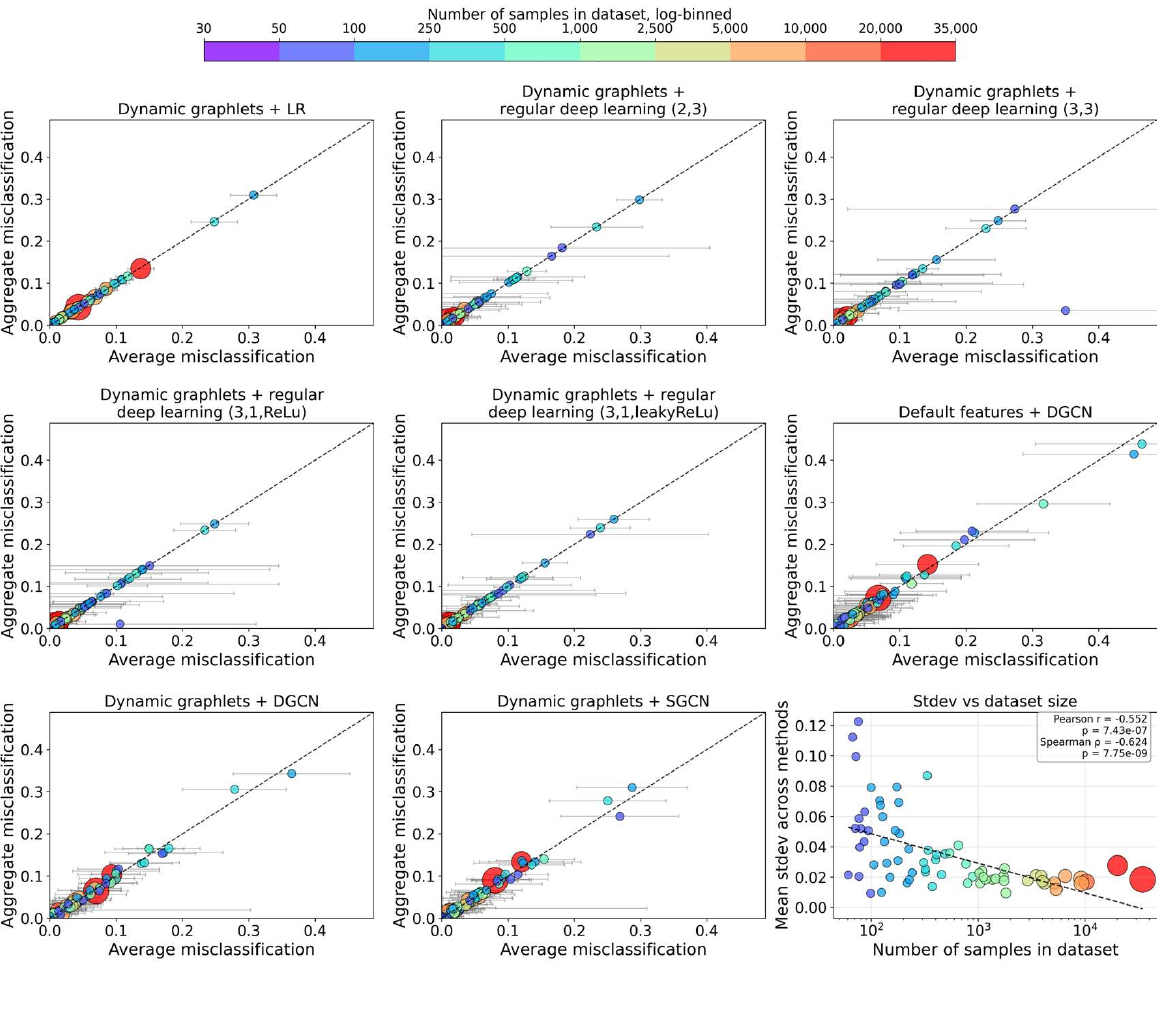}%
    \end{center}
    \vspace{-1cm}
    \caption{\textbf{All panels except bottom right:} The correlations between aggregate vs. average misclassification rates for each of the 72 datasets (dots in a given panel) for each of the eight dynamic PSN method evaluated in our study (corresponding to the eight panels). A standard deviation is drawn for each dot with respect to its average misclassification rate. These results indicate close-to-perfect correlation between average and aggregate misclassification rates. Because some of the datasets have large standard deviations for their average misclassification rates, in the bottom right panel, we perform an additional analysis, as follows. \textbf{The bottom right panel:}  The correlation between the dataset size (i.e. the number of samples in a dataset, $x$-axis) and the standard deviation (stdev) for a given dataset averaged over all eight methods. This result indicates a significantly strong negative correlation, i.e. that it is typically the smaller datasets that have large standard deviations. In each of the nine figure panels (including bottom right), a dot is a dataset, and its size and color correspond to the number of domains in that dataset.}
    \label{fig:sup_scatter}
\end{figure*}

Our discussion thus far has focused on reporting results for \emph{aggregate} misclassification rates. We have verified that even if we instead considered misclassification rates \emph{averaged} over the five cross-validation folds, the results would remain  almost identical, both qualitatively and quantitatively, because the aggregate and average misclassification rates correlate almost perfectly (Supplementary Fig. \ref{fig:sup_scatter}).

\section{Discussion and concluding remarks} \label{sect:conclution}

In this study, we investigate whether DL improves upon traditional ML in the task of \emph{dynamic PSN-based} PSC. We acknowledge that there exist other approaches that are not dynamic PSN-based, that could be used for PSC, and that might be competitive to dynamic PSNs. Comparison of dynamic PSNs to such approaches is not the focus of our study; dynamic PSNs have already been shown to be superior to static PSNs in the PSC task \cite{newaz2022multi}, and static PSNs have already been shown to be competitive to sequence and direct (non-network) 3D structural approaches in the same PSC task \cite{newaz2020network}, as well as in the related yet distinct tasks of protein structure comparison \cite{faisal2017grafene} and protein function prediction \cite{gligorijevic2021structure} (Section \ref{sect:introduction} and Fig. \ref{fig:overview}(a)). Instead, our focus in this paper is to examine in the context of a proven powerful existing PSC approach, namely dynamic PSNs, whether DL improves upon traditional ML.

We find that DL does not guarantee gain, and traditional ML holds its value. Similar was shown for protein function prediction: a (static) PSN-based GCN approach outperformed non-PSN approaches in that task \cite{gligorijevic2021structure}. Those same authors then found that (static) graphlet-based analyses outperformed GCN-based analyses in terms of function prediction accuracy \cite{berenburgyoutube1}. Nonetheless, the literature comparing traditional ML vs. DL is sparse \cite{zitnik2024current}. By examining whether DL adds value, especially if a powerful traditional ML approach exists for a given task (as it does for PSC), we add to the sparse literature. 

In the PSC task, dynamic PSNs and dynamic graphlets seem to be the primary driver of the impressive  performance of the considered traditional ML and DL approaches. Likely because of these drivers, even LR  already achieves low misclassification rates, e.g. of below 0.1 on $65/72=90\%$ of the datasets.
Consequently, there might not exist much room for DL to improve. Yet, DL does improve on some datasets, although it shows worse performance on others (see the ``absolute rank 1 (no ties)'' results in Fig. \ref{fig:results_main}(a)). 
For example, considering the $72-65=7$ datasets on which traditional ML achieves misclassification rates of 0.1 or higher, at least one considered DL approach achieves rates below 0.1 on $3/7=43\%$ of them. Note that we checked in our original dynamic PSN study \cite{newaz2022multi} whether the datasets where dynamic graphlets + LR (traditional ML) performed well vs. not so well were in any way ``biased'', e.g. towards any of the four hierarchy levels of CATH or SCOPe.  We could not detect any such ``bias'' \cite{newaz2022multi}. 

A natural question emerges: how to improve PSC from dynamic PSNs, be it via traditional ML or DL, on datasets where neither work well? For example, for the CATH-1.20.5 dataset, all approaches have misclassification rates over 0.25 (Fig. \ref{fig:results_over_all} and Supplementary Fig. \ref{fig:sup_fig_6_alt}). One option could be as follows. If 3D structural data were available on the actual intermediates along protein folding pathways, this would allow for capturing with dynamic PSNs the true folding dynamics rather than the ``proxy'' intermediates extracted from native 3D structures. This, in turn, would likely mean improved results from analyzing such dynamic PSNs in any task, including PSC. Unfortunately, such data are limited \cite{wells2026unavailability}. Another, more realistic option may be to design a hybrid approach taking the best from the different learning paradigms, per discussion in our recent work \cite{zitnik2024current}.

PSC is a task for which an effective traditional ML approach exists, i.e. even LR suffices given the dynamic PSN representations of 3D structures. Because of this, our PSC methodological and evaluation framework may be a valuable benchmark of future DL approaches for analyzing protein sequences or 3D structures, to ensure that they offer improvement compared to dynamic PSNs under LR, before they are then applied to downstream tasks such as protein function prediction. In fact, the development of improved benchmarks for the PSC task is an active effort \cite{Chen2026protein}. Thus, our study adds to this effort.

Speaking of protein function prediction, given the promise of static graphlets in that task (per the above discussion), it might be worth exploring function prediction from \emph{dynamic} PSNs. Certainly of our future interest is evaluation of dynamic PSN approach(es) against the most recent and powerful non-network approaches for analyzing protein sequences or structures, including both PSC approaches as well as protein structure comparison approaches such as Foldseek, in a consistent task such as protein function prediction.

In tasks where a traditional ML approach might not exist that is as effective as traditional ML is in the PSC task, especially where feature engineering becomes a bottleneck or where interactions are highly nonlinear, DL may provide substantial gains over traditional ML \cite{dai2025survey}. For predictive tasks on graphs, recent work on graph representation learning highlights that graph neural networks and transformers can model complex relational and multimodal biomedical data while enabling generalizability across predictive tasks and domains with minimal retraining \cite{feng2025comprehensive,johnson2024graph}. Another promising avenue for broad cross-domain generalization are the emerging graph foundation models \cite{wang2025graph}.
Such advanced DL could be crucial to translating scientific findings into real-world impact.

\section*{Acknowledgments}

We thank an undergraduate Brendan Haidinger for helping reproduce our results, and Dr. Monisha Ghosh for supporting her Ph.D. advisee Francis Gatsi's participation in this work.

\bibliographystyle{abbrv} 

\bibliography{bibliography}

\newpage

\section*{Supplementary Information}\label{sect:supplement}
\beginsupplement

\section{\textcolor{black}{Supplementary methods}}\label{sect:supporting_methods}

    \subsection{Data}\label{sect:sup_data}

Here, we add more details about the data. All datasets were derived from a large pool of experimentally determined protein 3D structures that were obtained and cleaned by K. Newaz et al. (2022) \cite{newaz2022multi} as follows. 145,219 entries, with crystallographic resolution $\leq 3\mathring{A}$, corresponding to 408,404 unique protein chains were obtained from PDB. Structural domains within these chains were identified using CATH and SCOPe annotations. To prevent inflated PSC performance due to sequence redundancy, a filtering step was applied to keep for further consideration only protein chains sharing less than 90\% pairwise sequence identity with any other chain in the set. This yielded 35,131 sequence-non-redundant chains containing 60,434 CATH and 25,864 SCOPe domains. To ensure sufficient statistical power for supervised classification, only those structural classes with at least 30 domains in a given class were kept. Further, small domains with fewer than 30 amino acids and ``noisy'' domains having disconnected PSNs were removed from consideration. In the end, this resulted in the 72 datasets spanning a total of 34,630 CATH and 9,329 SCOPe protein domains.

    \subsection{\textcolor{black}{Classification frameworks}}\label{sect:sup_classification_frameworks}

        \subsubsection{The regular DL architecture and its variants}\label{sect:sup_CNN_LSTM}

Here, we add more details about our considered regular DL variants, thus complementing the information in the main paper. 

Recall that the ``default'' regular DL variant that we consider is dynamic graphlets + (2 CNN, 3 LSTM). We use this variant because our lab's previous collaborative work \cite{guo2019weighted} showed that a regular DL design having two CNN layers and three LSTM layers was effective for the task of PSC using \textit{static} PSNs. In fact, at that time, two CNN layers were chosen over more CNN layers as the best of all evaluated options \cite{guo2019weighted}. Nonetheless, in our current study, because we are dealing with dynamic rather than static PSNs, to test the effect of the number of CNN layers, we do consider a larger number of CNN layers as well, namely three. We vary the number of LSTM layers in a similar fashion, considering up to three LSTM layers. In total, we consider four variants of regular DL, as described in the main paper. In theory, one could consider more CNN or LSTM layers. However, especially when it comes to the number of LSTM layers, as our analyses in this study show, increasing the number of layers can lead to significant increase in runtime. 

When it comes to parameter choice within a given regular DL variant, recall that for a given dataset containing $i$ protein domains where the total number of unique structural classes over all $i$ domains is $c$, each domain is represented by its corresponding non-zero dGDVM matrix of size $n \times 211$, where $n$ denotes the number of nodes in the dynamic PSN and 211 corresponds to the number of non-zero graphlet orbits. 
At the start of any of the regular DL variants, the non-zero dGDVM is processed by a given number of stacked CNN layers (i.e. two or three layers). The first CNN layer contains 56 kernels of size $5 \times 211$. Each kernel spans all 211 non-zero graphlets while covering five consecutive rows of the dGDVM, allowing the CNN to learn patterns among small groups of neighboring rows while spanning all graphlet orbits. During convolution, each kernel slides across the matrix along the row dimension (i.e. stride), ensuring that all possible five-row neighborhoods in the feature matrix are examined. Multiple kernels allow the network to learn different types of structural patterns from the same input. The second CNN layer contains 96 kernels of the same size, enabling the model to learn a larger set of feature detectors from the representations produced by the first CNN layer. In variants containing a third CNN layer (i.e. dynamic graphlets + (3 CNN, 3 LSTM), dynamic graphlets + (3 CNN, 1 LSTM), and dynamic graphlets + (3 CNN, 1 LSTM, LeakyReLU)), that third layer contains 128 kernels of the same size. For any of the CNN layers, a stride of 1 along the row dimension is applied. 
Because the LSTM layers learn from the ordered sequence of CNN-derived features, padding is used with the CNN layers to keep the input and output lengths the same. Padding does this by adding artificial elements, typically zeros, to the input matrix so that CNN filters can be applied at the boundaries without reducing the output size.
In addition, ReLU activation function (or LeakyReLU for the dynamic graphlets + (3 CNN, 1 LSTM, LeakyReLU) variant) is applied at the end of every CNN layer to introduce nonlinearity. 
A dropout with rate 0.3 is also applied as a regularization mechanism to mitigate overfitting -- by randomly deactivating a subset of neurons during training. After the CNN layers, for a PSN, the resulting feature matrix has dimensions $n \times 96$ for variants with two CNN layers or $n \times 128$ for variants with three CNN layers. For all feature matrices, they are passed to a recurrent component consisting of stacked bidirectional LSTM layers (having one or three layers, depending on the variant). A given LSTM layer has hidden size of 64 per each direction (with output at each step having a dimension of 64$+$64=128). The forget gate bias is initialized to 1.0, and dropout with rate of 0.3 is applied between LSTM layers (when applicable). The final LSTM layer (being the first or third layer, depending on the variant) produces a sequence of hidden states. The hidden states of the forward and backward directions are concatenated to produce a single 128-dimensional vector of the protein domain. This representation is passed through a fully connected layer and subsequently through a classification layer whose output dimension equals $c$, i.e. the number of unique structural classes over all $i$ domains in a given dataset. A SoftMax activation function produces a probability distribution over the $c$ classes, and the structural class assigned to (i.e. predicted for) a given domain is assigned as the class with the highest probability score out all of all c probability scores.

As with the LR framework, all training, testing, and hyperparameter tuning are conducted under stratified five-fold cross-validation (Section~\ref{sect:evaluation} in the main paper). All four regular DL variants are trained using cross-entropy loss for multi-class classification. Optimization is performed using the Adam optimizer with learning rate of $10^{-4}$. Gradient clipping with threshold of 3 is applied to reduce the risk of gradient explosion during training. For each dataset, models are trained for up to 100 epochs. To reduce the possibility that DL performance is limited by either incomplete training or overfitting, we monitor both training loss and validation loss across epochs, while using validation loss as the criterion for training control and selecting model parameters (i.e. the learned weights and biases of all CNN, LSTM, and fully connected layers). Specifically, we employ early stopping where, if validation loss does not improve within a fixed patience window of 20 epochs, training is terminated and the model parameters from the epoch with the lowest validation loss are retained. In addition, if validation loss does not improve for 10 consecutive epochs, the learning rate is automatically reduced (i.e. smaller parameter-update steps in subsequent epochs). This allows training to continue more cautiously when validation performance has plateaued, while early stopping terminates training if validation performance does not improve after a longer patience window. During training, the model parameters are saved whenever validation loss reaches a new minimum. After training terminates, we retain the saved parameters from the epoch with the lowest validation loss, rather than the parameters from the final epoch. This avoids retaining parameters from later epochs in which validation performance may have worsened, while still allowing training to continue long enough to determine whether validation performance can improve after a plateau.

We use TensorFlow/Keras to implement the CNN+LSTM variants. We use the scikit-learn library in Python for utilities such as splitting/evaluation.

        \subsubsection{Graph-based DL architecture and its variants}\label{sect:sup_GCN}

\paragraph{DGCN variant.} DGCN requires the input of a dynamic PSN and its corresponding node feature matrix. The following goes into detail on the preparation of both of these input types for downstream use in the DGCN.

\begin{itemize}
    \item (For a dynamic PSN) recall that each dynamic PSN consists of a set of $s$ network snapshots, where each snapshot is represented by an adjacency matrix. Because the number of amino acids for a dynamic PSN can differ across protein domains and across datasets, its $s$ adjacency matrices can vary in size. Because of this variability in size, we apply node padding to each snapshot of a dynamic PSN so that each adjacency matrix has size $p \times p$, where $p \times p$ corresponds to the dimensions of the final snapshot in a PSN. Node masking is also applied to distinguish real nodes from artificial padded nodes by maintaining a binary mask over node positions. This mask ensures that padded nodes are ignored in model computations, so they do not contribute to learned node embeddings, attention weights, or graph-level representations. 
    \item (For node features) a node feature matrix is initialized, padded, and masked with the same dimensionality ($p \times p$) as the padded adjacency matrices. Depending on the experimental setting, these features matrices can either padded dynamic graphlets (non-zero dGDVMs) or randomly initialized (`default'') features -- where feature values are sampled from a zero-centered distribution with variance scaled according to standard initialization schemes (i.e, He initialization).
\end{itemize}

Both the dynamic PSN and the node feature matrix are then passed to four stages in the DGCN variant: (i) spatial graph convolution, (ii) attention-based graph pooling, (iii) temporal modeling via LSTM, and (iv) classification. In stage (i), for a dynamic PSN and its corresponding node feature matrix, graph convolutions update each node’s feature by aggregating information from its neighboring nodes; with multiple GCN layers, information can also be incorporated from increasingly larger multi-hop neighborhoods. In stage (ii), attention-based graph pooling converts each PSN snapshot into a single graph-level embedding by assigning different importance weights to different nodes and summarizing their node representations. In stage (iii), the LSTM processes the ordered sequence of graph-level embeddings across PSN snapshots, allowing the model to combine information from earlier and later snapshots of the dynamic PSN. Finally, in stage (iv), the resulting features from stage (iii) are passed to a classifier, which outputs predicted class probabilities for the task of PSC. Below, we go into detail on each of the four stages: 

\ul{(i) Spatial graph convolution.} For a dynamic PSN, each padded adjacency matrix $A$ (corresponding to a PSN snapshot), initialized with its corresponding padded feature matrix $F$, is processed using two stacked GCN layers. GCN layers aggregate information from neighboring nodes in an adjacency matrix in order to learn representations that capture both local residue interactions and higher-order structural patterns. Two layers are used to allow information to propagate across multi-hop neighborhoods while maintaining a moderate model size and limiting overfitting. Both layers use a hidden dimensionality of size $64$; the first layer maps node features from $p \rightarrow 64$, and the second maps from $64 \rightarrow 64$. This dimensionality provides sufficient capacity to learn informative node embeddings while keeping the number of parameters manageable. Each layer performs symmetric adjacency normalization, 
\[ \hat{A} = F^{-1/2}(A + I)F^{-1/2}, \] 
where $I$ is the identity matrix and $F$ is the padded feature matrix. This normalization stabilizes training by preventing feature values from growing excessively during neighborhood aggregation. Node embeddings are then updated using 
\[ H^{(l+1)} = \hat{A}H^{(l)}W^{(l)} + b^{(l)}, \] 
where $H$ denotes the node embeddings, $W$ the learnable weight matrix, $l$ is the layer number, and $b$ the bias. Layer normalization and ReLU activation are applied after each GCN layer to stabilize optimization and introduce nonlinearity. Dropout rate of $0.3$ is also applied to reduce overfitting. The spatial convolution stage contains 26,944 learnable parameters and produces a node embedding matrix of size $p \times 64$ for each snapshot. 

\ul{(ii) Attention-based graph pooling.} Because downstream components of the DGCN framework require fixed-dimensional graph representations, node embeddings are aggregated into a single graph-level vector using attention pooling; attention pooling allows the method to learn which residues (nodes) contribute most strongly to the structural representation of the protein (dynamic PSN). For each node embedding, a two-layer MLP computes an importance score 
\[ e_i = W_2 \tanh(W_1 h_i + b_1) + b_2. \] 
The MLP consists of a linear layer ($64 \rightarrow 64$), Tanh activation, and a linear layer ($64 \rightarrow 1$), yielding 4,225 learnable parameters. The first linear layer preserves the dimensionality of the node embeddings while enabling the method to learn nonlinear transformations of the feature space. A Tanh activation function is used to introduce nonlinearity while producing bounded outputs, which can help stabilize the attention weights computation. The final linear layer maps the transformed node representation to a single scalar importance score, allowing the method to learn relative weights for nodes when constructing the graph-level embedding. Attention weights are computed via 
\[ \alpha_i = \frac{\exp(e_i)}{\sum_j \exp(e_j)}, \] 
and the graph-level embedding is obtained by 
\[ g_t = \sum_i \alpha_i h_i, \] 
resulting in a 64-dimensional embedding vector for each snapshot. 

\ul{(iii) Temporal modeling via LSTM.} Next, the set of snapshot embedding vectors forms a tensor of shape $[T,B,64]$, where $T$ denotes the number of snapshots and $B$ denotes batch size. To capture structural changes across snapshots, these embeddings are processed using a single-layer bidirectional LSTM with hidden size 64 per direction. We use an LSTM component for similar reasons described in  \ref{sect:sup_CNN_LSTM} and because it allows the method to capture temporal context from both earlier and later snapshots. For an input size of 64 and hidden size 32, the BiLSTM contains 12,416 learnable parameters. The output tensor has shape $[T,B,128]$. For each feature tensor input into the LSTM, the hidden states corresponding to the last valid timestep (determined by the padding mask) are extracted from both directions and concatenated, yielding a fixed-length embedding vector of size 128. This vector summarizes the temporal evolution of the PSN. 

\ul{(iv) Classification.} Lastly, the final feature vector is passed to a classification head consisting of a linear layer ($128 \rightarrow 32$), ReLU activation, dropout rate of $0.3$, and a final linear layer ($32 \rightarrow c$), where $c$ is the number of protein structural classes in a dataset. The intermediate projection reduces dimensionality while enabling the network to learn a compact embedding prior to classification. A SoftMax activation produces class probabilities. All together, these four stages define the DGCN framework employed in our study.

\paragraph{SGCN variant.}
To assess whether explicitly modeling of temporal dynamics improves PSC performance, we also evaluate a static graph convolutional network (SGCN) variant. In this model, only the final PSN snapshot (and its corresponding adjacency matrix and feature matrix) is used. The SGCN employs the same spatial graph convolution architecture as the DGCN (stage (i) above) and the same classification head as the DGCN (stage (iv) above). However, unlike in the DGCN, attention-based pooling (stage (ii) above) is not used in the SGCN, and instead a standard pooling component is used to produce graph embeddings. In addition, no temporal modeling stage (stage (iii) above) is applied in the SGCN. In this way, the SGCN serves as a control that isolates the contribution of temporal modeling, enabling us to determine whether incorporating dynamic PSN information provides benefits over learning solely from a single structural snapshot. 

For the training of both the DGCN and SGCN variant, we use cross-entropy loss optimized with the Adam optimizer (learning rate $10^{-3}$), which provides stable and efficient optimization through adaptive learning rates. Training is performed with batch size 1 for 100 epochs. A `ReduceLROnPlateau'' scheduler (factor$=0.5$, patience$=5$) is used to reduce the learning rate when validation performance plateaus, improving convergence. Note that evaluation and hyperparameter tuning for both the SGCN and DGCN variant follow the same five-fold cross-validation procedure used in the CNN+LSTM and LR frameworks. Both graph-based variant are implemented using the PyTorch library in Python.

\clearpage

\section{Supplementary figures}\label{sect:supporting_figs}

\clearpage

\clearpage

\clearpage

\clearpage

\clearpage

\begin{figure*}[ht]
    \begin{center}
            \includegraphics[
                width=1\textwidth,
                trim= 1cm 11cm 1cm 0cm,
                clip
            ]{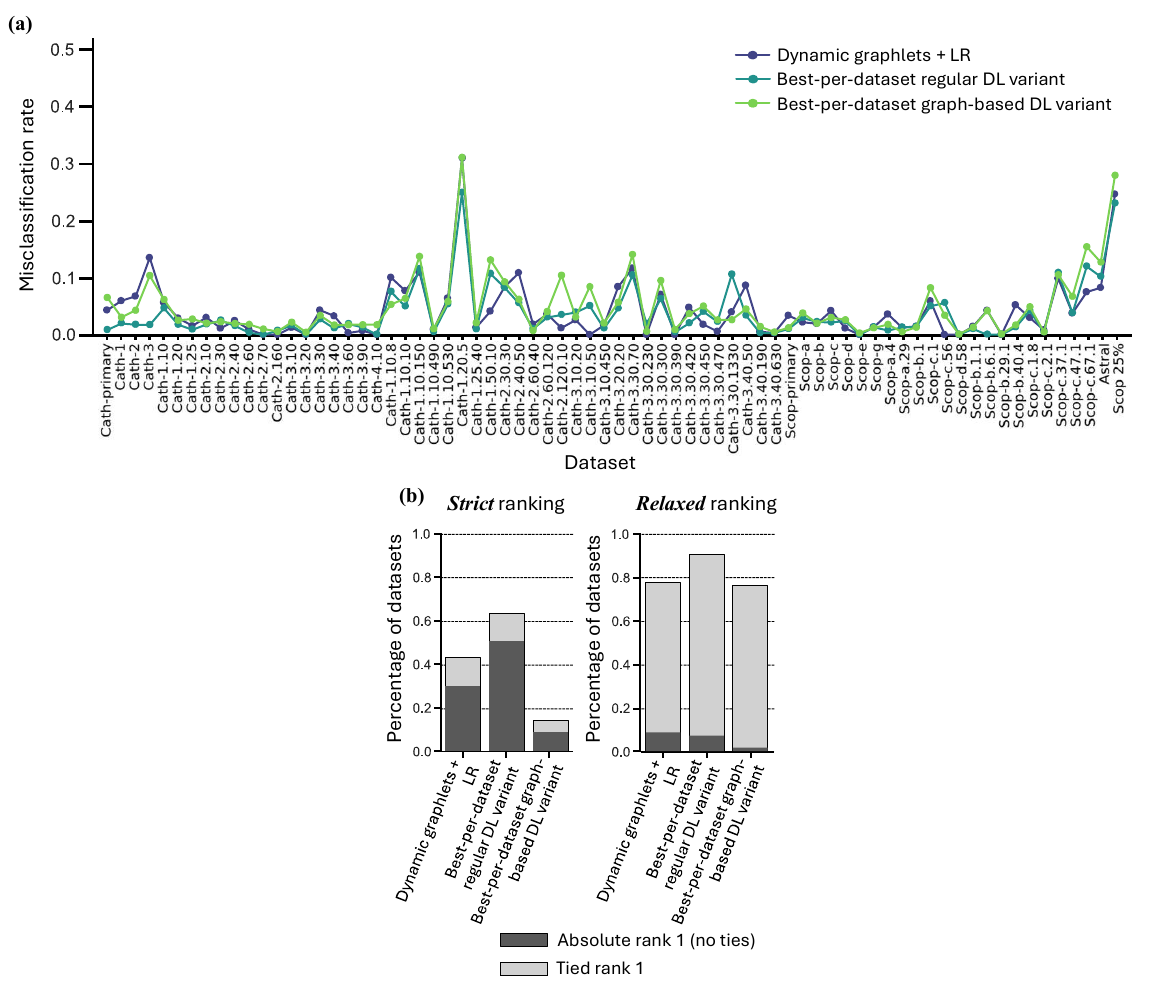}%
    \end{center}
    \caption{Comparison of traditional ML, regular DL, and graph-based DL in the \emph{best-per-dataset} analysis on 72 datasets in the task of dynamic PSN-based PSC. Specifically, on each dataset, the dynamic graphlets + LR approach, the best-on-that-dataset regular DL variant, and the best-on-that-dataset graph-based DL variant  are compared. \textbf{(a)} Per-dataset misclassification rates; recall that we report rates aggregated over all test folds of the 5-fold cross-validation. \textbf{(b)} Left: The percentage of all 72 datasets in which a given method is the best (rank 1), or is tied as the best (within 0\% absolute difference aka strict ranking). Right: The percentage of all 72 datasets in which a given method is the best (rank 1), or is close-to-tied as the best (within 2\% absolute difference aka relaxed ranking). Note that for the \emph{best-per-dataset} analysis we do not report runtime, because the choice of the best variant within a DL paradigm varies across datasets.}
    \label{fig:sup_fig_6}
\end{figure*}

\clearpage

\begin{figure*}[ht]
    \begin{center}
            \includegraphics[
                width=1\textwidth,
                trim= 0.5cm 17.5cm 5.5cm 0cm,
                clip
            ]{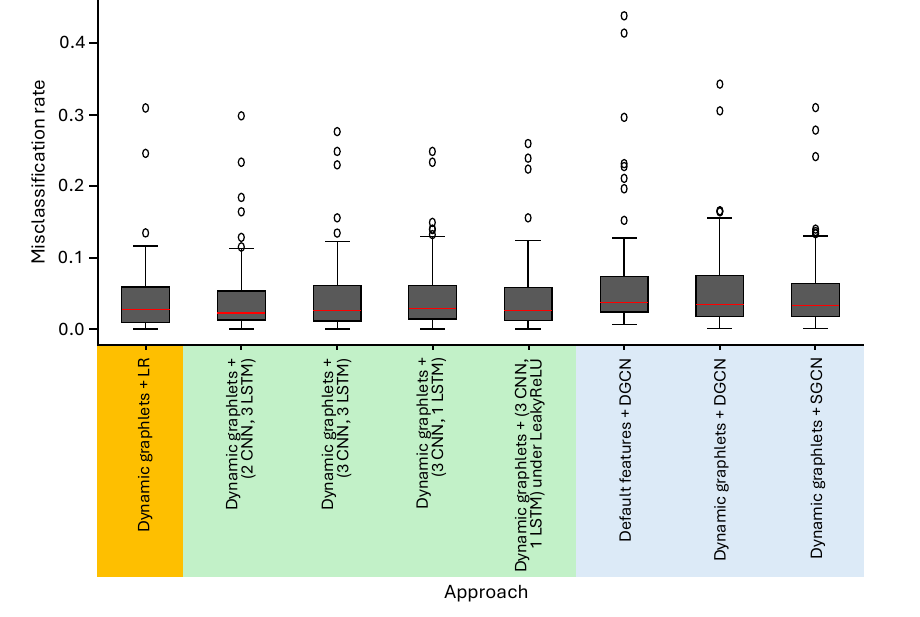}%
    \end{center}
    \caption{The distribution of a method’s misclassification rates over all 72 datasets. The red line is the distribution mean, and outliers are shown as circles.}
    \label{fig:sup_fig_7}
\end{figure*}

\clearpage

\begin{figure*}[ht]
    \begin{center}
            \includegraphics[
                width=1\textwidth,
                trim= 1.5cm 19cm 2cm 0cm,
                clip
            ]{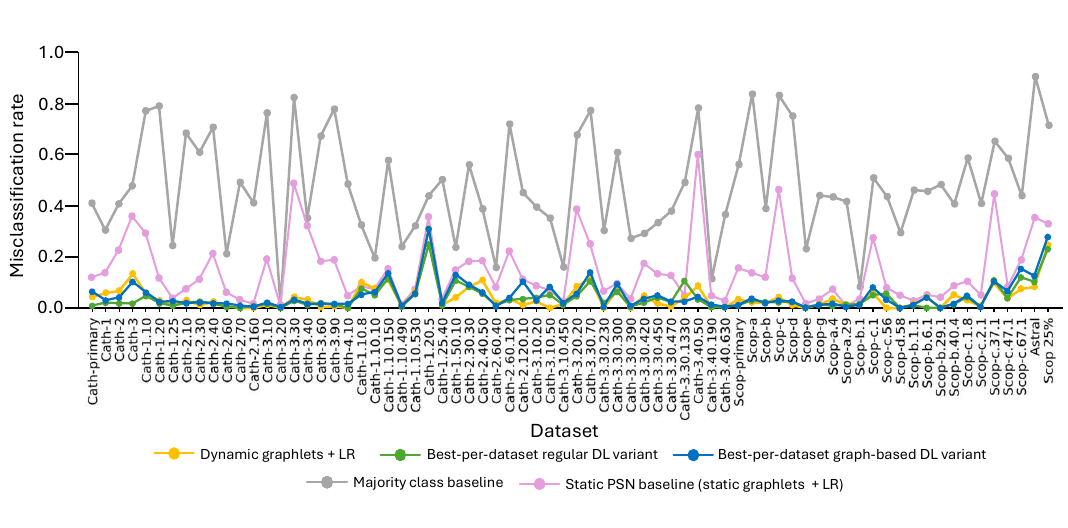}%
    \end{center}
    \caption{Per-dataset misclassification rates in the \emph{best-per-dataset} analysis; recall that we report rates aggregated over all test folds of the 5-fold cross-validation. Note: this is the same figure as  Supplementary Fig. \ref{fig:sup_fig_6}(a) but with two additional baseline approaches. The key comparison is between traditional ML (dynamic graphlets + LR), the best per-dataset variant within the regular DL paradigm, and the best per-dataset variant within the graph-based DL paradigm. To illustrate the power of these approaches that are all based on dynamic PSNs, we add results for two baselines that do not use dynamic PSNs. One is majority class, which is what the misclassification rate would be if all protein domains in a given dataset were predicted  to have the largest of all classes in the dataset. The other one is based on static PSNs \cite{newaz2020network,newaz2022multi}; this is the static counterpart of dynamic graphlets + LR.}
    \label{fig:sup_fig_6_alt}
\end{figure*}

\clearpage

\clearpage

\end{document}